%% file: main.tex
\documentclass[%
 aip,
 amsmath,amssymb,
 reprint,%
]{revtex4-1}

\usepackage{graphicx}
\usepackage{bm}
\usepackage[utf8]{inputenc}
\usepackage[T1]{fontenc}
\usepackage{mathptmx}
\usepackage{etoolbox}
\usepackage{svg}
\usepackage{verbatim}
\usepackage{soul}

\usepackage{enumitem}
\usepackage{xcolor}
\usepackage[hidelinks]{hyperref}
\usepackage[capitalise]{cleveref}
\usepackage{caption}
\usepackage{subcaption}
\usepackage{ragged2e}
\usepackage[cal=cm]{mathalfa}

\begin{document}

\title{Learning Beyond Experience: Generalizing to Unseen State Space with Reservoir Computing}
\author{Declan A. Norton}
\email[Correspondence email address: ]{nortonde@umd.edu}
\affiliation{Department of Physics, University of Maryland, College Park, Maryland 20742, USA\looseness=-1}
\author{Yuanzhao Zhang}
\affiliation{Santa Fe Institute, Santa Fe, New Mexico 87501, USA\looseness=-1}
\author{Michelle Girvan}
%\email[Correspondence email address: ]{girvan@umd.edu}
\affiliation{Department of Physics, University of Maryland, College Park, Maryland 20742, USA\looseness=-1}
\affiliation{Santa Fe Institute, Santa Fe, New Mexico 87501, USA\looseness=-1}
\affiliation{Institute for Research in Electronics and Applied Physics, University of Maryland, College Park, Maryland 20742, USA\looseness=-1}
\affiliation{Institute for Physical Science and Technology, University of Maryland, College Park, Maryland 20742, USA\looseness=-1}

\input{sections/Abstract.tex}
\pacs{}% insert suggested PACS numbers in braces on next line
\maketitle
\input{sections/Lead_Paragraph.tex}
\input{sections/Intro.tex}
\input{sections/Methods.tex}
\input{sections/Results.tex}
\input{sections/Discussion.tex}
\input{sections/Supplemental_Statement.tex}
\input{sections/Acknowledgements.tex}
\input{sections/Author_Declarations.tex}
\input{sections/Bibliography.tex}
\input{sections/Appendix.tex}
\input{sections/Supplemental.tex}

\end{document}

%% file: sections/Abstract.tex
\begin{abstract}
Machine learning techniques offer an effective approach to modeling dynamical systems solely from observed data. However, without explicit structural priors -- built-in assumptions about the underlying dynamics -- these techniques typically struggle to generalize to aspects of the dynamics that are poorly represented in the training data. Here, we demonstrate that reservoir computing -- a simple, efficient, and versatile machine learning framework often used for data-driven modeling of dynamical systems -- can generalize to unexplored regions of state space without explicit structural priors. First, we describe a multiple-trajectory training scheme for reservoir computers that supports training across a collection of disjoint time series, enabling effective use of available training data. Then, applying this training scheme to multistable dynamical systems, we show that RCs trained on trajectories from a single basin of attraction can achieve out-of-domain generalization by capturing system behavior in entirely unobserved basins.
\end{abstract}

%% file: sections/Lead_Paragraph.tex
\begin{quotation}
    A plethora of machine learning (ML) techniques have shown impressive success in modeling complex dynamical systems from observed data alone -- a challenge of broad practical importance, e.g., in climate science, public health, economics, ecology, and neuroscience. However, these so-called `black-box' models, which do not incorporate guidance based on known or suspected features of the system, often require extensive training data that cover the range of possible system behaviors. When the training domain is incomplete, they often fail outside of it, whereas models that incorporate knowledge of the system's properties may still succeed. Here, we show that reservoir computing -- a black-box ML approach commonly used to model dynamical systems -- can overcome this limitation in important settings. By studying systems that can evolve toward multiple distinct long-term behaviors, we show that reservoir computers can make accurate predictions about behaviors they have never seen before, even when trained on limited data.
\end{quotation}

%% file: sections/Intro.tex
\section{Introduction} \label{sec:Intro}

Data-driven methods for modeling dynamical systems are essential in applications where a system of interest is complex or poorly understood and no sufficient knowledge-based (i.e., mathematical or physical) model is available. A number of machine learning (ML) techniques have proven effective for this purpose.\cite{Brunton2019_Book,Han2021_Deep_TS_Prediction_Review} When choosing among these ML approaches, however, one typically faces a trade-off between data-efficiency and model flexibility.

Unlike black-box approaches, methods that exploit partial knowledge of the system of interest through a structural prior, i.e., an explicit assumption or constraint about the system’s form, are often capable of generalizing to regions of the system's state space not sampled in the training data (out-of-domain generalization),\cite{Zhang2023_Catch22,Goring2024_OutOfDomain,Gauthier2022_LearningUnseenCoexistingAttractors,Yu2024_PhysicsGuidedDL} making them data-efficient and robust. Some such methods constrain the functional form of the ML model, e.g., sparse identification of nonlinear dynamics\cite{Brunton2016_SINDy,Rudy2017_PDESindy} (SINDy) and next generation reservoir computing\cite{Gauthier2021_NGRC} (NGRC), while others combine ML models with imperfect knowledge-based components in hybrid configurations.\cite{Pathak_Hybrid,Arcomano2022_AtmHybrid,Chepuri2024_RC_NGRC_Hybrid} If the inductive bias conferred to the model by its structural prior is inconsistent with the system of interest, however, the model performance can deteriorate substantially.\cite{Zhang2023_Catch22}

On the other hand, black-box models that incorporate no explicit priors (but often have implicit inductive biases, which may be subtle\cite{Vardi2023_ImplicitBias,Ribeiro2021_DoubleDescentModelingDynamics}) can be expressive enough to model diverse systems of interest, making them highly flexible. When presented with data outside of their training context, however, these models cannot rely on system-informed constraints to help them generalize. Thus, they typically perform poorly in regions of state space not well sampled by their training data.\cite{Zhang2023_Catch22,Goring2024_OutOfDomain,Gauthier2022_LearningUnseenCoexistingAttractors,Rohm2021_UnseenAttractors_RC,Du2024_PowerSystemBasinsML,Yu2024_PhysicsGuidedDL} Black-box models that fall into this class include a broad set of techniques based on artificial neural networks, from recurrent architectures -- e.g., reservoir computers (RCs), long short-term memory networks (LSTMs), and gated recurrent units (GRUs) -- to feedforward methods such as neural ODEs.

Here, contrary to widely held assumptions, we demonstrate that reservoir computers (RCs) -- a simple and efficient ML framework commonly used to learn and predict dynamical systems from observed time series\cite{JaegarHaas,ESANN_Res_Overview,Sun2024_ESN_Review,Lukosevicius2009_RC_Review} -- can generalize to unexplored regions of state space in many relevant settings, even without explicit structural priors to guide their behavior.
 
The simplicity of RCs makes them versatile, and they have been employed for a wide range of purposes: inferring unmeasured system variables from time series data,\cite{Lu_and_Pathak} forecasting dynamics of extended networks\cite{Srinivasan2022_ParallelRC_for_Networks} or  spatiotemporal systems,\cite{Pathak2018_SpatioTemporalChaosRC} separation of chaotic signals,\cite{Krishnagopal2020_Sep_of_Chaotic_Signals} inferring network links,\cite{Banerjee2021_RCLinkInference} and more.\cite{Lukosevicius2009_RC_Review,Tanaka_RC_Review_2019,Bollt_RC_VAR} As with other black-box forecasting approaches, however, previous studies using RCs have focused mostly on monostable systems -- those that exhibit a single stable long-term behavior, which is confined to a particular region of state space. For these systems, it suffices to train an RC on a single long time series that samples the relevant region well. Here, to test RCs' out-of-domain generalization ability, we apply reservoir computing to the challenging problem of basin prediction in `multistable' systems -- that is, systems in which each trajectory evolves towards one of multiple distinct long-term behaviors, i.e., `attractors,' each confined to a different region of state space and having a corresponding `basin of attraction' containing all initial conditions whose trajectories converge to that attractor.

Because basins of attraction are non-overlapping, multistable systems present a natural setting to test RCs' ability to generalize to unexplored regions of state space. Such systems also arise frequently in important scenarios\cite{Wagemakers2025_BasinZoo} (e.g., neuroscience,\cite{Izhikevich2006_DynSysNeuroscience} gene regulatory networks,\cite{Rand2021_GeneRegulatoryDynamics} cell differentiation and pattern formation,\cite{Corson2017_Drosophila} electrical grids,\cite{Menck2014_PowerGridStability,Du2024_PowerSystemBasinsML} and financial markets\cite{Cavalli2016_MultistabilityMarketGames}) and are often too complex to confidently construct ML models with suitable structural priors. They remain underexplored, however, using black-box ML approaches.

In this paper, we describe a scheme to train RCs on a collection of disjoint time series, allowing for more flexible and exhaustive use of available data. This multiple-trajectory training has previously been applied to multi-task learning, where the goal is to train a single RC across multiple dynamical systems, each of which exhibit different dynamics.\cite{Norton2025_METAFORS,Kong2021_MLSystemCollapse,Panahi2024_AdaptableRC,Kong2024_IndexBasedRC,Kim2020_RNNChaoticMemories,Lu2020_IGS} Here, we leverage the scheme's flexibility to improve sampling of the state space in multistable dynamical systems with short-lived transients. Then, we utilize the multistability of these systems to test when RCs can generalize from their training data to capture system dynamics in unexplored regions of state space. Specifically, we show that an RC trained on trajectories from a single basin of attraction can recover the dynamics in other unexplored basins, capturing even fractal-like basin structures and unseen chaotic attractors.

%% file: sections/Methods.tex
\section{Challenges in Predicting Multistable Systems}\label{sec:Challenge}

While basin prediction from a short initial time series is fundamentally a challenge related to `climate replication' (predicting the long-term statistics) of dynamical systems, which has been well studied with RCs in monostable dynamical systems,\cite{Lu2018_Attractor_Reconstruction,Patel2021_NonstationaryRC,Norton2025_METAFORS,Panahi2025_CriticalTransitions,Panahi2024_AdaptableRC} it differs from the traditionally studied monostable scenario in ways that make it substantially more challenging. Here, we highlight these challenges in the context of reservoir computing.

\begin{figure}[t]
	\centering
	\includegraphics[width=\linewidth]{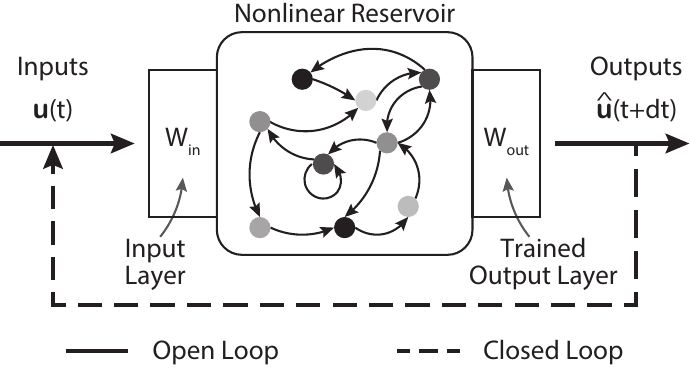}
	\caption{\justifying\textbf{A Reservoir Computer for Time Series Prediction.}}
	\label{fig:Res_Diagram}
\end{figure}

Reservoir computers (RCs) predict the evolution of a system with state $\boldsymbol{x}$ whose dynamics are governed by
\begin{equation}\label{eq:Generic_TrueDS}
    \frac{d\boldsymbol{x}}{dt}=\boldsymbol{f}(\boldsymbol{x}),
\end{equation}
by constructing an auxiliary dynamical system -- the `reservoir system' -- with `reservoir state' $\boldsymbol{r}$ that evolves according to its own dynamical equation,
\begin{equation}\label{eq:OpenLoop_AuxDS}
    \frac{d\boldsymbol{r}}{dt}=\boldsymbol{F}\left[\boldsymbol{r}(t),\boldsymbol{u}(t)\right],
\end{equation}
where ${\boldsymbol{u}(t)}$ is a driving signal. To train an RC for forecasting, we drive the reservoir system with an observed time series that is a function of the state of the true system, ${\boldsymbol{u}(t)=\boldsymbol{g}(\boldsymbol{x}(t))}$, in the `open-loop mode' (\cref{fig:Res_Diagram}). Then we choose a linear readout matrix or `output layer', $W_{out}$, such that $\boldsymbol{u}(t)$ can be approximated through a linear projection of the auxiliary state:
\begin{equation}\label{eq:Output_Goal}
   \hat{\boldsymbol{u}}(t)= W_{out}\boldsymbol{r}(t) \approx \boldsymbol{u}(t).
\end{equation}
Once we have a suitable output layer, the RC can evolve as an autonomous dynamical system, using its own output as the driving signal in the `closed-loop mode' (\cref{fig:Res_Diagram}), to mimic the system of interest:
\begin{equation}\label{eq:ClosedLoop_AuxDS}
    \frac{d\boldsymbol{r}}{dt}=F[\boldsymbol{r}(t),\hat{\boldsymbol{u}}(t)]=F[\boldsymbol{r}(t),W_{out}\boldsymbol{r}(t)].
\end{equation}

Typical approaches for predicting monostable dynamical systems with RCs assume that a single long training series, $\boldsymbol{u}(t)$, that evolves along the system's single stable attractor -- a manifold $M_{sys}$ -- provides data that are well sampled on $M_{sys}$. So long as certain conditions are satisfied,\cite{Jaeger2001_ESP,Lukosevicius_2012,Cucchi2022_Hands_On_RC,Platt2021_PredictiveGS,Platt2022_RC_for_Complex_Forecasting_Review} the reservoir system, when driven by the training series $\boldsymbol{u}(t)$, will evolve along some corresponding manifold, $M_{res}$, in the state space of the reservoir once a transient response of the reservoir has passed.\cite{Lu2018_Attractor_Reconstruction,Platt2021_PredictiveGS,Platt2022_RC_for_Complex_Forecasting_Review} A well-trained output layer thus represents a mapping from the manifold $M_{res}$ to the manifold $M_{sys}$ and encodes the dynamics of the true system on this attracting manifold. The output layer may not, however, accurately represent the dynamics of the true system in regions of state space that were not explored in training; i.e., regions that are separated from the manifolds $M_{res}$ and $M_{sys}$.

Multistable dynamical systems, which have more than one attracting manifold, thus present two challenges to forecasting with reservoir computers. (1)~It is often difficult to obtain training time series that sufficiently sample the state space of multistable dynamical systems. Since basins of attraction are necessarily non-overlapping, a single training series cannot sample more than one of the attracting manifolds. Moreover, the transient behaviors of trajectories that have not yet reached their attractors are hard to sample -- the transients do not lie on any of the attracting manifolds, and are also frequently short-lived. (2)~Basins of attraction often have complex, intertwined boundaries, so that a system's final state depends sensitively on its initial condition.\cite{Grebogi1983_FinalStateSensitivity} In such cases, a relatively small prediction error at one time step can push a trajectory from the correct basin of attraction to an incorrect basin of attraction, making climate replication much more challenging.

In the next section, we provide a detailed description of our reservoir computing implementation and of the multi-trajectory training scheme we use to facilitate more exhaustive use of disjoint training time series, allowing for better sampling of transient dynamics.

\section{Training a Reservoir Computer on Multiple Trajectories}\label{sec:Methods}

To construct and train reservoir computers (RCs) for our experiments, we use some of the same implementations as used in previous work by DN, MG, and collaborators,\cite{Norton2025_METAFORS} built on the \textit{rescompy} python package.\cite{rescompy} Accordingly, portions of our RC implementation and its description are adapted from that earlier work.\cite{Norton2025_METAFORS}

The central component of an RC is a recurrent neural network, `the reservoir', whose nodes, indexed by $i$, have associated continuous-valued, time-dependent activation levels $r_i(t)$. The activations of all $N_r$ nodes in the reservoir constitute the reservoir state, ${\boldsymbol{r}(t)=[r_1(t),...,r_{N_r}(t)]^T}$, which evolves in response to an input signal according to a dynamical equation with a fixed discrete time step, $\Delta t$:
\begin{equation}
\label{eq:Open_Res_Update}
\begin{aligned}
    \boldsymbol{r}(t+\Delta t)= & (1-\lambda)\boldsymbol{r}(t) \\ & +\lambda\tanh(W_r\boldsymbol{r}(t)+W_{in}\boldsymbol{u}(t)+\boldsymbol{b}),
\end{aligned}
\end{equation}
where the $\tanh()$ function is applied element-wise. The input weight matrix, $W_{in}$, couples the $N_{in}$-dimensional input ${\boldsymbol{u}(t)}$ to the reservoir nodes. The directed and weighted adjacency matrix, $W_r$, specifies the strength and sign of interactions between each pair of nodes, and a random vector of biases, $\boldsymbol{b}$, breaks symmetries in the nodes' dynamics. We say that the reservoir has `memory' if its state ${\boldsymbol{r}(t + \Delta t)}$ depends not only on the most recent input, ${\boldsymbol{u}(t)}$, but also (recursively) on previous inputs, ${\boldsymbol{u}(t-m\Delta t)}$ for ${m>0}$ -- i.e., if ${(1-\lambda)}$ and/or the matrix $W_r$ are nonzero. The leakage rate, $\lambda$, thus influences the time-scale on which the reservoir state evolves, and, consequently, the duration of its memory.

The adjacency matrix, $W_r$, of each reservoir is a sparse, random, directed network with mean degree $\langle d\rangle$ and probability of connection between each pair of nodes given by ${\langle d\rangle/N_r}$. We assign non-zero elements of $W_r$ random values from a uniform distribution ${\mathcal{U}[-1, 1]}$ and then normalize this randomly generated matrix such that its spectral radius (eigenvalue of largest absolute value) has some desired value, $\rho$. To generate the dense input matrix, $W_{in}$, and the bias vector, $\boldsymbol{b}$, we choose each entry from the uniform distributions ${\mathcal{U}[-\sigma,\sigma]}$ and ${\mathcal{U}[-\psi,\psi]}$, respectively. We call $\sigma$ the input strength range and $\psi$ the bias strength range.

To train an RC for a forecasting task, we choose its output layer, of dimension ${N_{in}\times N_r}$, so that at every time step over a set of ${N_{train}}$ training signals, ${\boldsymbol{u}^{(i)},\,i=1,...,N_{train}}$, which we standardize such that each component has mean zero and maximum absolute value one (as measured across the union of all training signals), the RC's output closely matches its input at the next time step:
\begin{equation}
    \boldsymbol{u}^{(i)}(t+\Delta t)\approx W_{out}\boldsymbol{r}^{(i)}(t+\Delta t).
    \label{eq:Training_Targets}
\end{equation}
The internal parameters of the reservoir ($W_r$, $W_{in}$, $\boldsymbol{b}$, and $\lambda$) are set prior to training and remain unaltered thereafter. To calculate the output layer, we add white noise to the input time series in order to promote stable predictions\cite{Wikner2024_LMNT}
\begin{equation}
    \tilde{u}_j^{(i)}(t)=u_j^{(i)}(t)+\mathcal{N}\left(0,\eta\, \mathrm{RMS}[u_j]\right),
    \label{eq:Training_Noise}
\end{equation}
where ${\mathrm{RMS}[u_j]}$ is the root-mean-square amplitude of the $j^\mathrm{th}$ component of the inputs calculated over all training time series, ${\mathcal{N}\left(0,\xi\right)}$ draws a random sample from a Gaussian distribution with mean zero and standard deviation $\xi$, and $\eta$ is a small constant -- the `noise amplitude.' We then drive the reservoir with the noisy training signals in the open-loop mode (\cref{eq:OpenLoop_AuxDS,eq:Open_Res_Update,fig:Res_Diagram}) and minimize the ridge-regression cost function:
\begin{equation}\sum_{i=1}^{N_{train}}\sum_{n=N_{trans}}^{N_{i}-1}\frac{\| W_{out}\boldsymbol{r}^{(i)}(n\Delta t)-\tilde{\boldsymbol{u}}^{(i)}(n\Delta t)\|^2}{N_{fit}}+\alpha\|W_{out}\|^2,
    \label{eq:Ridge_Cost}
\end{equation}
where $N_i$ is the number of (evenly-spaced) data points in the $i^\mathrm{th}$ signal (i.e., it has duration ${(N_{i}-1)\Delta t}$), the scalar $\alpha$ is a (Tikhonov\cite{Tikhonov1995}) regularization parameter which prevents over-fitting, $\|\ \|$ denotes the Euclidean ($L^2$) norm, and ${N_{fit}=\sum_{i=1}^{N_{train}}N_{i}-N_{train}(N_{trans}+1)}$ is the number of input/output pairs used for fitting. Importantly, we discard the first $N_{trans}$ reservoir states and target outputs of \textit{each} training signal as a transient to allow the reservoir state to synchronize to each signal before fitting over the remaining time steps. (Note that \cref{eq:Ridge_Cost} reduces to the usual cost function for single-trajectory training when ${N_{train}=1}$.) The minimization problem, \cref{eq:Ridge_Cost}, has solution
\begin{equation}
    W_{out}=YR^T\left(RR^T+\alpha N_{fit}I\right)^{-1},
    \label{eq:Pred_Ridge}
\end{equation}
where $I$ is the identity matrix and $Y$ (${N_{in}\times N_{fit}}$) and $R$ (${N_r\times N_{fit}}$), respectively, are the target and reservoir state trajectories over the fitting periods.

We highlight that the dimensions of the matrices ${YR^T}$ 
and ${RR^T}$ are independent of the number of training signals, $N_{train}$. This fact is useful when $N_{fit}$ is large (either because we wish to train across a large number of input time series or because the time series are long) and storing the reservoir states in computer memory becomes a challenge. In such cases, we generate batches, ${b_i,\,i=1,...,N_b}$, of reservoir states, each of which is small enough to store, and calculate the total feature matrix ${RR^T}$ as the sum of the feature matrices of the batches, ${RR^T=\sum_{i=1}^{N_b}(RR^T)_{b_i}}$. Once we have calculated the feature matrix ${(RR^T)_{b_i}}$ for batch $b_i$, we can discard the reservoir states for that batch and move on to the next batch. Thus, we need store only one batch of reservoir states at a time. We can calculate ${YR^T}$ similarly.

Once the RC has been trained as described above, we use it to obtain predictions, $\hat{\boldsymbol{u}}(t)$, of the system of interest. During the prediction phase, the RC operates in closed-loop mode: at each time step, its input is set to its own output from the previous step, allowing it to evolve as an autonomous dynamical system. Used in this way, the RC is intended to mimic the behavior of the system of interest as in~\cref{eq:ClosedLoop_AuxDS}:
\begin{subequations}\begin{equation}
\begin{aligned}
    \boldsymbol{r}(t+\Delta t) = & (1-\lambda)\boldsymbol{r}(t) \\ & +\lambda\tanh(W_r\boldsymbol{r}(t)+W_{in}\hat{\boldsymbol{u}}(t)+\boldsymbol{b}),
\end{aligned}\label{eq:Closed_Loop_Res}\end{equation}\begin{equation}
    \hat{\boldsymbol{u}}(t+\Delta t)=W_{out}\boldsymbol{r}(t+\Delta t).\label{eq:Closed_Loop_u}
\end{equation}\label{eq:Closed_Loop_Updates}\end{subequations}

In typical applications, we wish to predict how the system of interest will evolve, having observed its recent behavior over some period. In this case, we naturally use these recent observations to initialize the forecast. Namely, we drive the reservoir in open-loop mode (\cref{eq:Open_Res_Update}) with a short `test signal,' $\boldsymbol{u}_{test}$, consisting of the available recent observations and then switch to the closed-loop mode (\cref{eq:Closed_Loop_Updates}) to forecast from the end of $\boldsymbol{u}_{test}$. The test signal enables the state of the auxiliary reservoir system to synchronize to the state of the underlying system of interest. In general, an appropriate test signal can be substantially shorter than would be sufficient to train an RC accurately. Hence, an RC that has been trained to accurately capture the dynamics of $\boldsymbol{u}(t)$, can be used to predict from a different initial condition by starting from a comparatively short test signal. The test signal should, however, be at least as long as the RC's memory to ensure that the memory is appropriately initialized. (A few recent studies have also proposed methods to `cold-start' forecasts with test signals that are even shorter than the RC's memory.\cite{Norton2025_METAFORS,Grigoryeva2024_ColdStartRC})

%% file: sections/Results.tex
\section{Results} \label{sec:Results}

We evaluate the ability of our RC setup to generalize to unexplored regions of state space using simulated data from four multistable dynamical systems: the Duffing system,\cite{Duffing_1918} a multi-well system with segregated basins, a magnetic pendulum system with fractal-like basins,\cite{Motter2013_DoublyTransient} and a multistable Lorenz-like system with coexisting chaotic attractors.\cite{Lu2004_MultistableLorenz} Previous studies\cite{Goring2024_OutOfDomain,Zhang2023_Catch22} have shown that learning the dynamics of even low-dimensional multistable systems can be challenging for typical RC approaches.

To evaluate the performance of our RC implementation in the context of dissipative systems with fixed-point attractors, we adopt a working definition for when a time series $\boldsymbol{u}(t)$ is said to converge to a fixed point $\boldsymbol{A}^i$, ${\boldsymbol{A}(\boldsymbol{u})=\boldsymbol{A}^i}$.  Specifically, if $\boldsymbol{A}^i$ is the nearest stable fixed point to the final point of the series, $\boldsymbol{u}(t_f)$, our convergence criteria require that
$\boldsymbol{u}(t)$ satisfies one of two additional conditions, depending on whether the system state is fully or partially measured. When $\boldsymbol{u}(t)$ contains full system state information at every time step, we require that the energy of the system at $t_f$, $E\left[\boldsymbol{u}(t_f)\right]$, is below the potential barrier, $E_0$, between the system's stable attractors: ${E\left[\boldsymbol{u}(t_f)\right]<E_0}$. When the full system state is not available and we cannot calculate the system energy, we instead require that the final $25$ points of ${\boldsymbol{u}(t)}$ are all within a threshold distance, $\varepsilon_c$, of the attracting fixed point $\boldsymbol{A}^i$.

Given a set of true system trajectories, $\{\boldsymbol{u}^{k}\ :\ k=1,...,N\}$, and corresponding predicted trajectories $\{\hat{\boldsymbol{u}}^{k}\ :\ k=1,...,N\}$, we thus approximate the true basin of attraction for the attractor $\boldsymbol{A}^i$ as the set of initial conditions whose trajectories converge to $\boldsymbol{A}^i$:
\begin{equation}
    B\left(\boldsymbol{A}^i\right)=\{\boldsymbol{u}^k(0)\ :\ \boldsymbol{A}(\boldsymbol{u}^k)=\boldsymbol{A}^i,\ k=1,...,N\}.
    \label{eq:True_Basins}
\end{equation}
Similarly, we estimate the predicted basin of attraction of $\boldsymbol{A}^i$ as 
\begin{equation}
    \hat{B}\left(\boldsymbol{A}^i\right)=\{\hat{\boldsymbol{u}}^k(0)\ :\ \boldsymbol{A}(\hat{\boldsymbol{u}}^k)=\boldsymbol{A}^i,\ k=1,...,N\}.
\label{eq:Predicted_Basins}
\end{equation}

\input{sections/Duffing_Results.tex}
\input{sections/Multi_Well_Results.tex}
\input{sections/Magnetic_Pendulum_Results.tex}
\input{sections/Lorenz_Results.tex}

%% file: sections/Duffing_Results.tex
\subsection{Intertwined Basins in the Duffing System}
\label{subsec:Duffing}

\begin{figure}[!th]
    \centering
    \includegraphics[width=.9\linewidth]{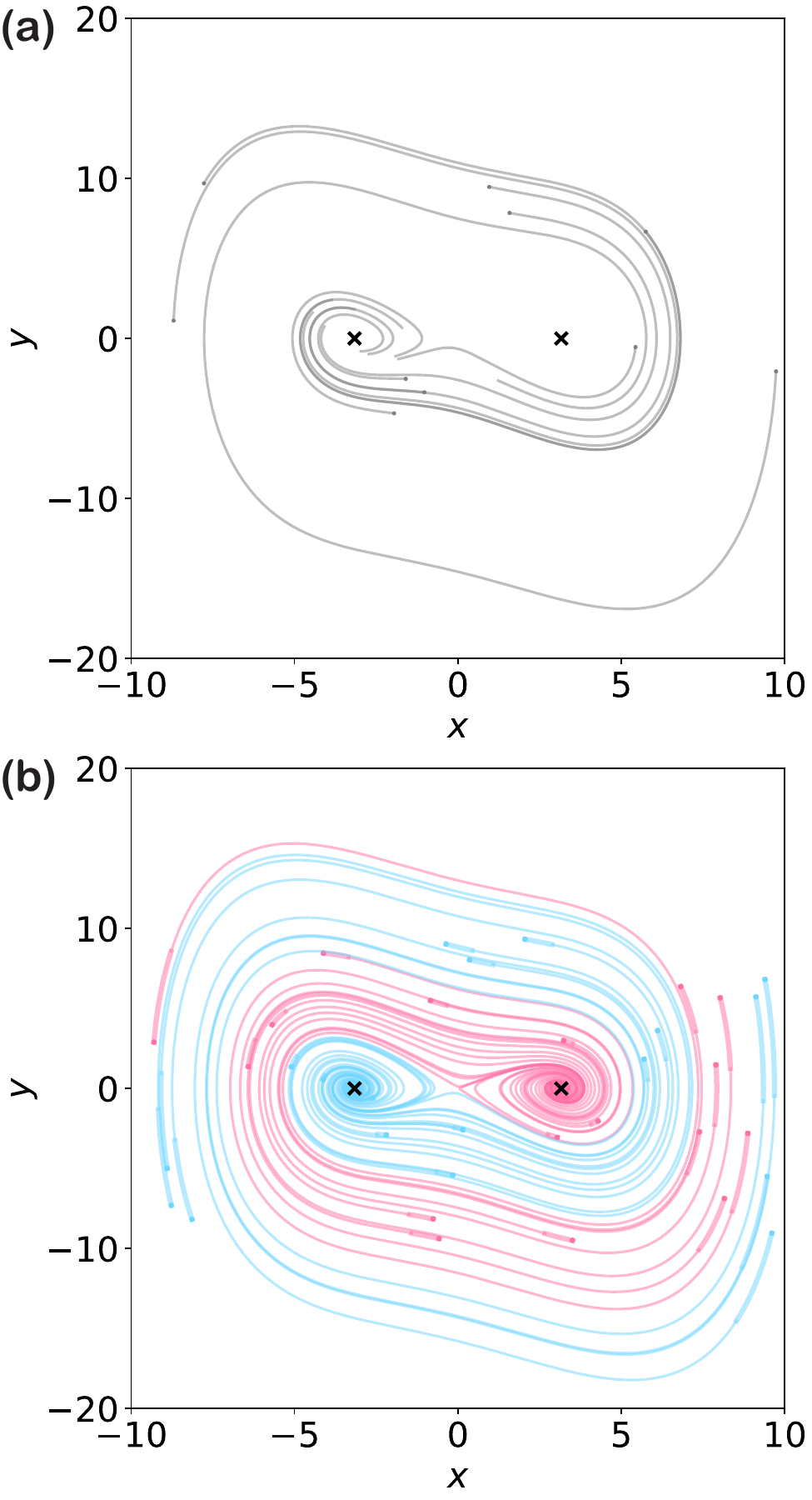}
    \caption{\justifying \textbf{RC identification of an unseen attractor in the Duffing system.} \textbf{(a)} We train an RC with ${N_r=200}$ nodes on ${N_{train}=10}$ fully-observed trajectories (gray lines) from one of the Duffing system's basins of attraction. \textbf{(b)} Then, we forecast the system's evolution from $36$ initial conditions (dots) and corresponding short test signals (thick lines, ${N_{test}=10}$ observations). Predictions are illustrated by thin lines. We color each trajectory according the true basin of its initial condition. The RC recovers system behavior in both the seen (blue) and unseen (pink) basins, and predicts the correct fixed point for all sample test signals.
    }
    \label{fig:Duffing_Phase_Space}
\end{figure}

The Duffing system\cite{Duffing_1918} models an oscillator moving under a nonlinear elastic force with linear damping. It is governed by a pair of coupled ordinary differential equations:
\begin{subequations}\label{eq:Duffing}
\begin{equation}\label{eq:Duffing_xdot}
\dot{x}=y,
\end{equation}
\begin{equation}\label{eq:Duffing_ydot}
\dot{y}=F_0+ay-bx-cx^3.
\end{equation}
\end{subequations}
With ${a=-1/2}$, ${b=-1}$, and ${c=1/10}$, the Duffing system is dissipative and multistable, with two attracting fixed points. When the system is unforced ($F_0=0$), the positions of these attractors are
\begin{equation}
    \boldsymbol{A}^\pm=\left(\pm\sqrt{-b/c},0\right)\approx\left(\pm3.16,0\right).
\end{equation}
There is also an unstable fixed point at the origin. Trajectories starting from almost any initial condition ${(x_0,y_0)}$ will thus converge to one of the attractors, $\boldsymbol{A}^\pm$. The initial conditions from which trajectories converge to $\boldsymbol{A}^+$ form the basin of attraction ${B(\boldsymbol{A}^+)}$ and the initial conditions from which trajectories converge to $\boldsymbol{A}^-$ form the basin of attraction ${B(\boldsymbol{A}^-)}$. Only trajectories that move along the stable manifold of the unstable fixed point do not converge to one of the two attractors. The corresponding set of initial conditions, which is of measure zero, forms the boundary between ${B(\boldsymbol{A}^+)}$ and ${B(\boldsymbol{A}^-)}$.

\begin{figure*}[!th]
    \centering
    \includegraphics[width=.95\linewidth]{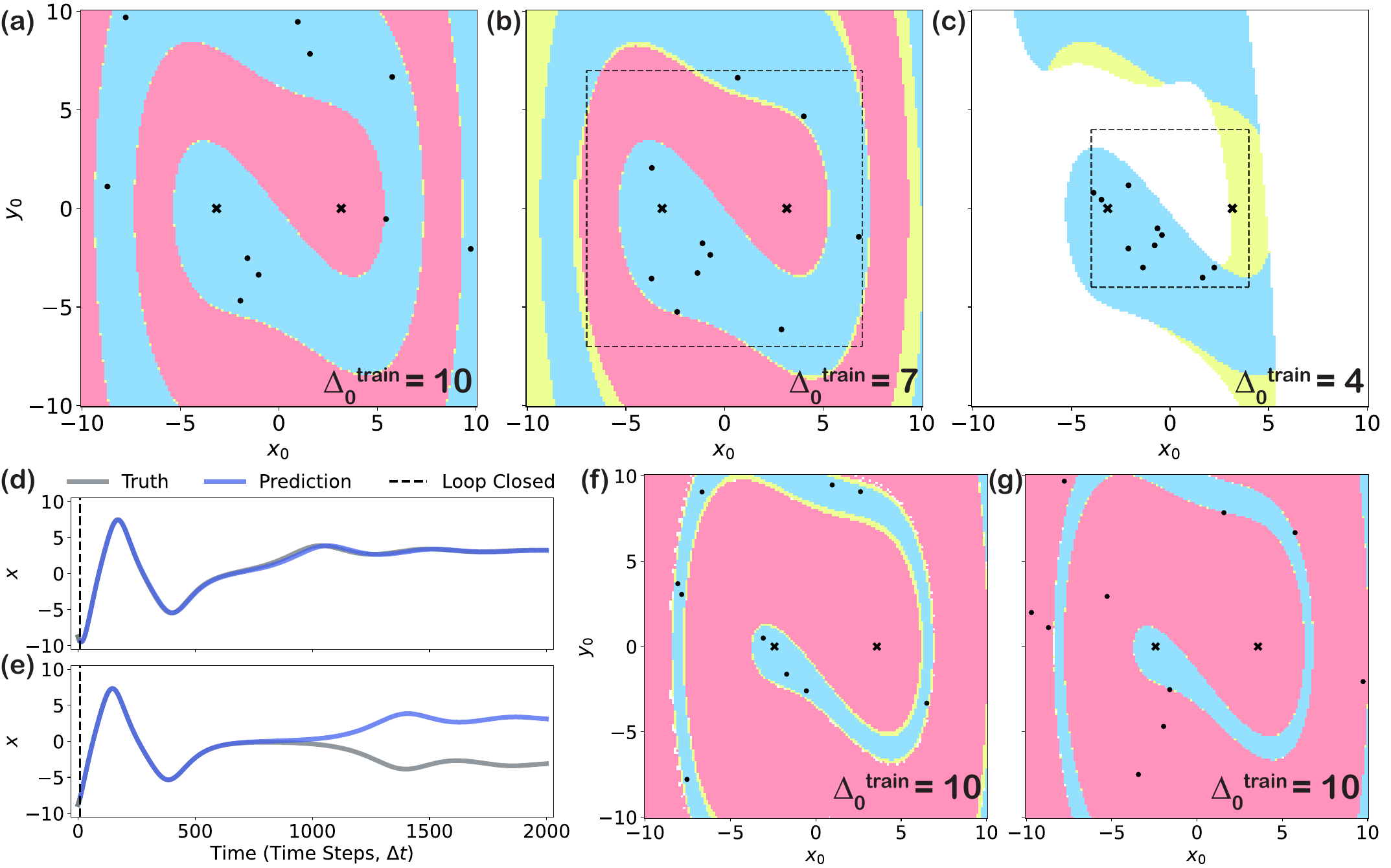}
    \caption{\justifying \textbf{RCs can generalize to unseen basins even with sparse and restricted training data.} \textbf{(a) to (e)} We train an RC of ${N_r=200}$ nodes on ${N_{train}=10}$ partially-observed trajectories ($x$ only) of the unforced Duffing system ($F_0=0$) from the basin of attraction $B(\boldsymbol{A}^-)$ (blue). Then we make predictions from short test signals (${N_{test}=10}$ observations each) from both basins of attraction. We draw the initial conditions, $(x_0,y_0)$, of the training signals from a random distribution that uniformly covers the intersection of $B(\boldsymbol{A}^-)$ with a square of side $2\Delta_0^{train}$ centered at the origin (dashed boxes), i.e. ${x_0\sim\mathcal{U}\left[-\Delta_0^{train},\Delta_0^{train}\right]}$ and ${y_0\sim\mathcal{U}\left[-\Delta_0^{train},\Delta_0^{train}\right]}$, subject to $(x_0,y_0)\in B(\boldsymbol{A}^-)$. \textbf{(a) to (c)}: The RC-predicted basins for different values of $\Delta_0^{train}$, from ${\Delta_0^{train}=10}$ (a) to ${\Delta_0^{train}=4}$ (c). The blue/pink initial conditions are those that the RC correctly predicts belong to the seen/unseen basin. The initial conditions of test signals for which the RC-predicted trajectory converges to the incorrect attractor are yellow. White initial conditions indicate that the predicted trajectory converges to a spurious attractor. Crosses mark the true fixed point attractors and dots mark the training initial conditions. We use $y_0$ for plotting purposes only; the RC has access to and then predicts only the $x$-component of the Duffing system. \textbf{(d) and (e)}: Example predictions for $\Delta_0^{train}=10$. The test signal ends at the vertical dashed line. \textbf{(f) and (g)} The RC still generalizes to unseen basins when a constant external forcing (${F_0=1}$) breaks the Duffing system's rotational symmetry, whether we train the RC on the smaller basin (f) or the larger basin (g).
    }
    \label{fig:Duffing_Basins}
\end{figure*}

We show in \cref{fig:Duffing_Phase_Space} that a reservoir computer (RC) trained on trajectories from only one basin of attraction, ${B(\boldsymbol{A}^-)}$, can capture the unforced Duffing system's dynamics in both basins of attraction and infer the existence of the other fixed-point attractor, which is unseen in the training data. 
Specifically, we train the RC on ${N_{train}=10}$ different trajectories, each of which converges to the fixed point $\boldsymbol{A}^-$ (\cref{fig:Duffing_Phase_Space}a). We then evaluate the trained RC on short \textit{test signals} with initial conditions sampled from both basins of attraction (\cref{fig:Duffing_Phase_Space}b). Each test signal consists of the first ${N_{test}=10}$ observations of the true system's evolution from a new initial condition, representing the kind of partial trajectory typically available in a prediction task. These test signals serve both to initialize the forecast and to allow the reservoir to synchronize to the new system state before entering the autonomous closed-loop mode. Even though the RC's training data explores only the basin of attraction of the fixed point $\boldsymbol{A}^-$, the RC infers the existence and location of the second attractor, $\boldsymbol{A}^+$, and correctly predicts that $\boldsymbol{A}^+$ is also a fixed point (and not, for example, a limit cycle).

In \cref{fig:Duffing_Basins}a-e, we likewise train an RC on $N_{train}=10$ trajectories from the basin of attraction $B(\boldsymbol{A}^-)$, but give the RC access to just the $x$-component of each trajectory, so that it receives only partial information of the system state at every time step. Then, we again forecast from test signals in both basins of attraction, each consisting of $N_{test}=10$ observations of the true system's evolution. The sample forecasts in \cref{fig:Duffing_Basins}d-e highlight the challenging nature of the basin prediction problem. The trajectories of the true system in both cases are almost identical until the system approaches the unstable fixed point at the origin around time step $500$. In \cref{fig:Duffing_Basins}d, the RC correctly predicts that the trajectory converges to the unseen fixed point, $\boldsymbol{A}^+$. In \cref{fig:Duffing_Basins}e, however, the true trajectory converges to the seen fixed point, $\boldsymbol{A}^-$, while the RC predicts that it converges to the unseen fixed point. As the system approaches the stable manifold of the unstable fixed point at the origin (in this case, it approaches the origin itself), the seemingly small prediction error is sufficient to push the autonomous RC system into the incorrect basin of attraction, and the predicted and true trajectories then separate rapidly.

In \cref{fig:Duffing_Basins}a we make predictions from test signals with initial conditions ${(x_0,y_0)}$ arranged in a ${150\times150}$ grid spanning ${-10\leq x_0\leq10}$ and ${-10\leq y_0\leq10}$, and plot the RC-predicted basin structure. As before, the RC has access to just the $x$-component of each test signal; we use $y_0$ for plotting purposes only. Test signals for which the RC correctly predicts that the system converges to the seen and the unseen attractors are colored blue and pink. Yellow indicates test signals for which the RC predicts the incorrect attractor. The ${N_{train}=10}$ initial conditions of the training signals, all from the basin ${B(\boldsymbol{A}^-)}$, are marked by black dots and the positions of the two attracting fixed points are marked by black crosses.

\cref{fig:Duffing_Basins}a demonstrates clearly that an RC trained on data from only one basin of attraction is able to generalize to the unexplored basin. The RC not only infers the existence of the unseen fixed point, but it also achieves high accuracy in predicting whether any given initial condition belongs to the basin $B(\boldsymbol{A}^-)$ or the basin $B(\boldsymbol{A}^+)$. Only near the basin boundary, where a trajectory's final state is most sensitive to perturbations, does the RC occasionally predict the incorrect basin.

\begin{figure}[!t]
    \centering
    \includegraphics[width=\linewidth]{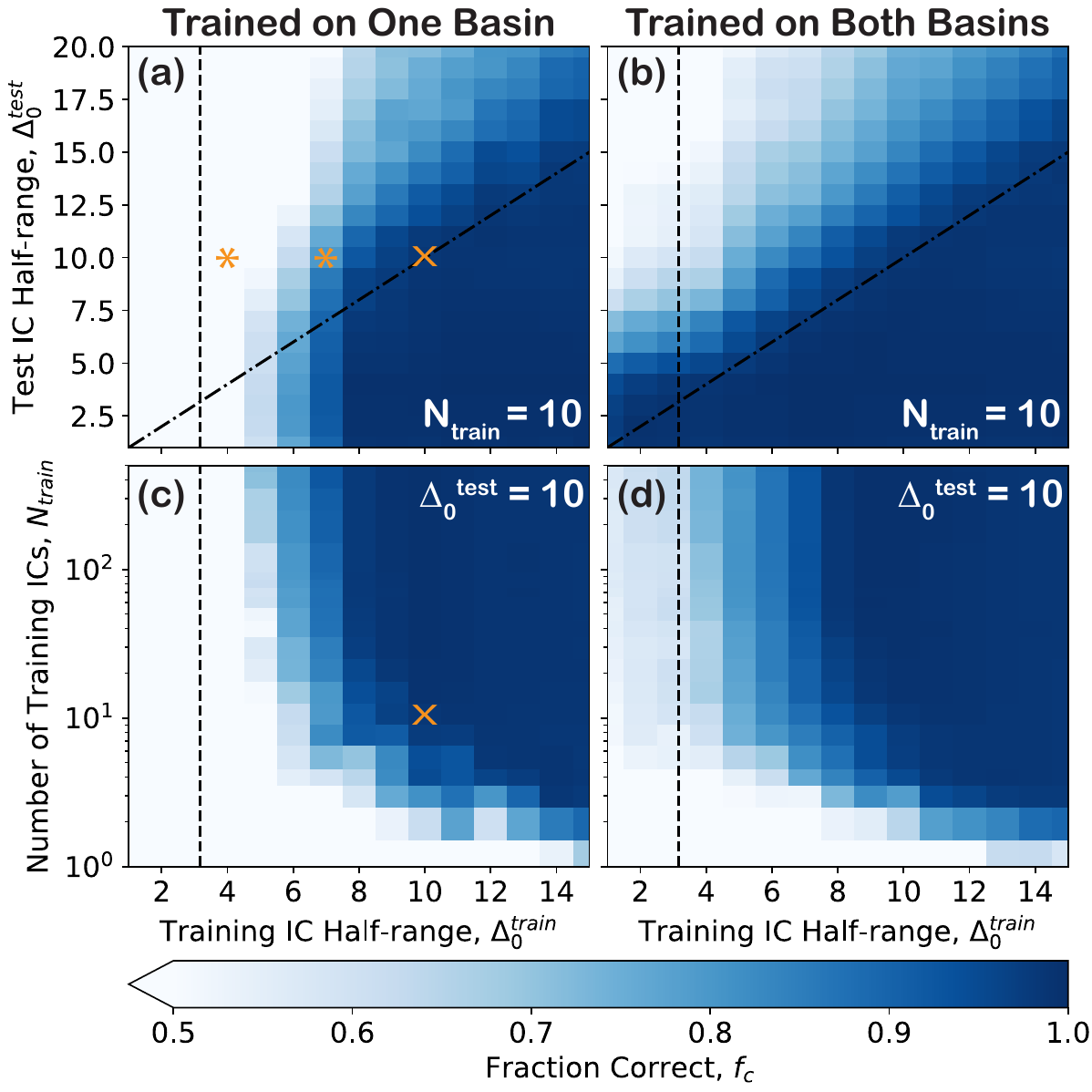}
    \caption{\justifying \textbf{In many scenarios, RC performance in unseen basins is comparable to performance in the training basin.} We train an RC of ${N_r=200}$ nodes to forecast partially-observed states of the Duffing system ($x$ only) and plot the fraction of short test signals for which the RC predicts the correct basin of attraction as we vary the number of training initial conditions, $N_{train}$, and the half-widths of the ranges of the training and test initial conditions, ${\Delta_0^{train}}$ and ${\Delta_0^{test}}$. \textbf{(a)~and~(c)}:~All training signals are from ${B(\boldsymbol{A}^-)}$. \textbf{(b)~and~(d)}:~Training signals are from both basins, ${B(\boldsymbol{A}^\pm)}$. In all cases, the training initial conditions are random, but the initial conditions of the test signals, belonging to both basins, form a ${50\times50}$ uniform grid. Each test signal consists of ${N_{test}=10}$ consecutive data points from the true system, starting from the specified initial condition. For all grid points, we plot the mean fraction correct calculated over ten random realizations of the RC's internal structure and of the training initial conditions. \textbf{(a)~and~(b):}~${N_{train}=10}$. \textbf{(c)~and~(d):}~${\Delta_0^{test}=10}$. Vertical lines mark the distance of the fixed points from the origin, $||\boldsymbol{A}^\pm||$, and diagonal lines mark ${\Delta_0^{test}=\Delta_0^{train}}$. Crosses and asterisks indicate the values of $N_{train}$, $\Delta_0^{train}$, and $\Delta_0^{test}$ that we use in \cref{fig:Duffing_Basins}a,d,e and in \cref{fig:Duffing_Basins}b-c, respectively}
    \label{fig:Duffing_Heatmaps}
\end{figure}

To further explore the ability of RCs to generalize, we investigate the effects of training data diversity, e.g., in terms of the range of initial conditions sampled. In \cref{fig:Duffing_Basins}b-c, we reduce the half-width, ${\Delta_0^{train}}$, of the range from which we select training initial conditions, while holding fixed the half-width of the test initial condition range, ${\Delta_0^{test}=10}$. Here, we see that for the RC to reliably predict system behavior in both basins, the initial conditions of the training trajectories must be sufficiently far from the fixed-point attractors. For ${\Delta_0^{train}=7}$ (\cref{fig:Duffing_Basins}b), the RC is still able to infer the existence of the unseen attractor and accurately reconstruct a large part of the unseen basin. However, for some initial conditions in the outer regions of the test grid, it predicts that the corresponding trajectories converge to the incorrect attractor (yellow). When the training range is reduced to ${\Delta_0^{train}=4}$ (\cref{fig:Duffing_Basins}c), the RC fails to identify the unseen fixed-point attractor $\boldsymbol{A}^+\approx(3.16,0)$. Trajectories starting in the white regions of \cref{fig:Duffing_Basins}c instead converge to spurious fixed-point attractors near $x\approx-8.4$ and $x\approx1.75$ (not shown) that are inconsistent with the Duffing system's true dynamics. In the supplementary material (\cref{fig:Full_Duffing_Trajectories}), we present a similar result with RCs trained on fully-observed states of the Duffing system and a similar spurious attractor is clearly visualized.

We demonstrate in \cref{fig:Duffing_Basins}f-g, that the RC does not rely on the Duffing system's rotational symmetry to generalize to the unexplored basin. Specifically, we break the symmetry by adding a constant external forcing, ${F_0=1}$, to \cref{eq:Duffing_ydot}. This change also moves the attracting fixed points to ${\boldsymbol{A}^-_F\approx\left(-2.42,0\right)}$ and ${\boldsymbol{A}^+_F\approx\left(3.58,0\right)}$. We then train the RC on ${N_{train}=10}$ partially-observed trajectories from a single basin of attraction and make predictions from short test signals (${N_{test}=10}$) in both basins of attraction, as before. The RC still recovers the system behavior in both basins, whether we train it on data from the smaller blue basin (panel f) or from the larger pink basin (panel g).

In \cref{fig:Duffing_Heatmaps}, we compare the performance of RCs trained on trajectories from only one basin of attraction of the unforced Duffing system, ${B(\boldsymbol{A}^-)}$, to that of RCs trained on trajectories from both basins. To quantify performance, we measure the fraction of trajectories for which the RC predicts the correct attractor,
\begin{equation}
\begin{split}
    f_c=&\frac{\text{Number of Correctly Predicted Basins}}{\text{Total Number of Predictions}}.
\end{split}
\end{equation}
At every grid point of the heatmaps in \cref{fig:Duffing_Heatmaps}, we plot the mean fraction correct, $f_c$, averaged over ten independent random draws of the RC's internal connections, $W_{in}$ and $W_{r}$, and of the initial conditions of the training trajectories.

In \cref{fig:Duffing_Heatmaps}a and b, we vary the half-ranges, ${\Delta_0^{train}}$ and ${\Delta_0^{test}}$, of both the training and test initial conditions. We note that that when ${\Delta_0^{train}}$ is sufficiently large ($\Delta_0^{train}\gtrsim7$) and the training time series sample the transient dynamics of the Duffing system far from its attractors, RCs trained on trajectories from only one basin of attraction capture the system's basin structure just as reliably as those trained on trajectories from both basins. The degradation in performance above the diagonal in both cases, however, demonstrates that RCs have difficulty extrapolating to regions of state space far from their training data -- even as those trained on trajectories from only one basin reliably generalize to the unexplored basin. Finally, the white region to the left of \cref{fig:Duffing_Heatmaps}a, where the fraction correct is no better than chance, highlights again that RCs trained on only one basin fail to generalize to the unexplored basin if their training signals do not sufficiently sample the transient dynamics of the Duffing system far from its attractors, as illustrated in \cref{fig:Duffing_Basins}. In contrast, RCs trained on data from both basins still perform well when $\Delta_0^{train}$ is small, so long as $\Delta_0^{test}$ is also small.

In \cref{fig:Duffing_Heatmaps}c and d, we vary the number of training trajectories, $N_{train}$, while holding the range of test initial conditions fixed at ${\Delta_0^{test}=10}$ (for consistency with \cref{fig:Duffing_Basins}, and because it is large enough that the RC must capture the Duffing system's transient dynamics well to offer good basin prediction). Here, we see that there is little or no generalization gap. That is, there is no substantial difference in accuracy between RCs trained on data from both basins and those trained on data from only one. In both cases, basin prediction is challenging if the training initial conditions are restricted to a narrow range, even if a large number of training trajectories is available. On the other hand, if the training trajectories are drawn from a wide range of initial conditions and sample well the Duffing system's transient dynamics, RCs can offer useful basin predictions with very few training trajectories.

We demonstrate in the supplementary material (\cref{fig:Duffing_Dynamical_Noise}) that RCs can still reliably generalize to an unseen basin of the unforced Duffing system when moderate stochastic forcing (dynamical/process noise) influences the evolution of the training trajectories.

%% file: sections/Multi_Well_Results.tex
\subsection{Segregated Basins in a Multi-well System}
\label{subsec:Multi_well}

In the Duffing system, the basins of attraction for the two fixed-point attractors are substantially intertwined. When the training initial conditions span a wide range of phase space, some of the resulting trajectories from the observed basin spend time close to regions belonging to the unseen basin of attraction, providing useful information for generalization. In this section, we investigate how well our RC setup can generalize in a system with segregated basins of attraction, where training trajectories may start near basin boundaries but quickly move away from them, making the extrapolation task potentially more challenging.

\begin{figure}[!th]
    \centering
    \includegraphics[width=\linewidth]{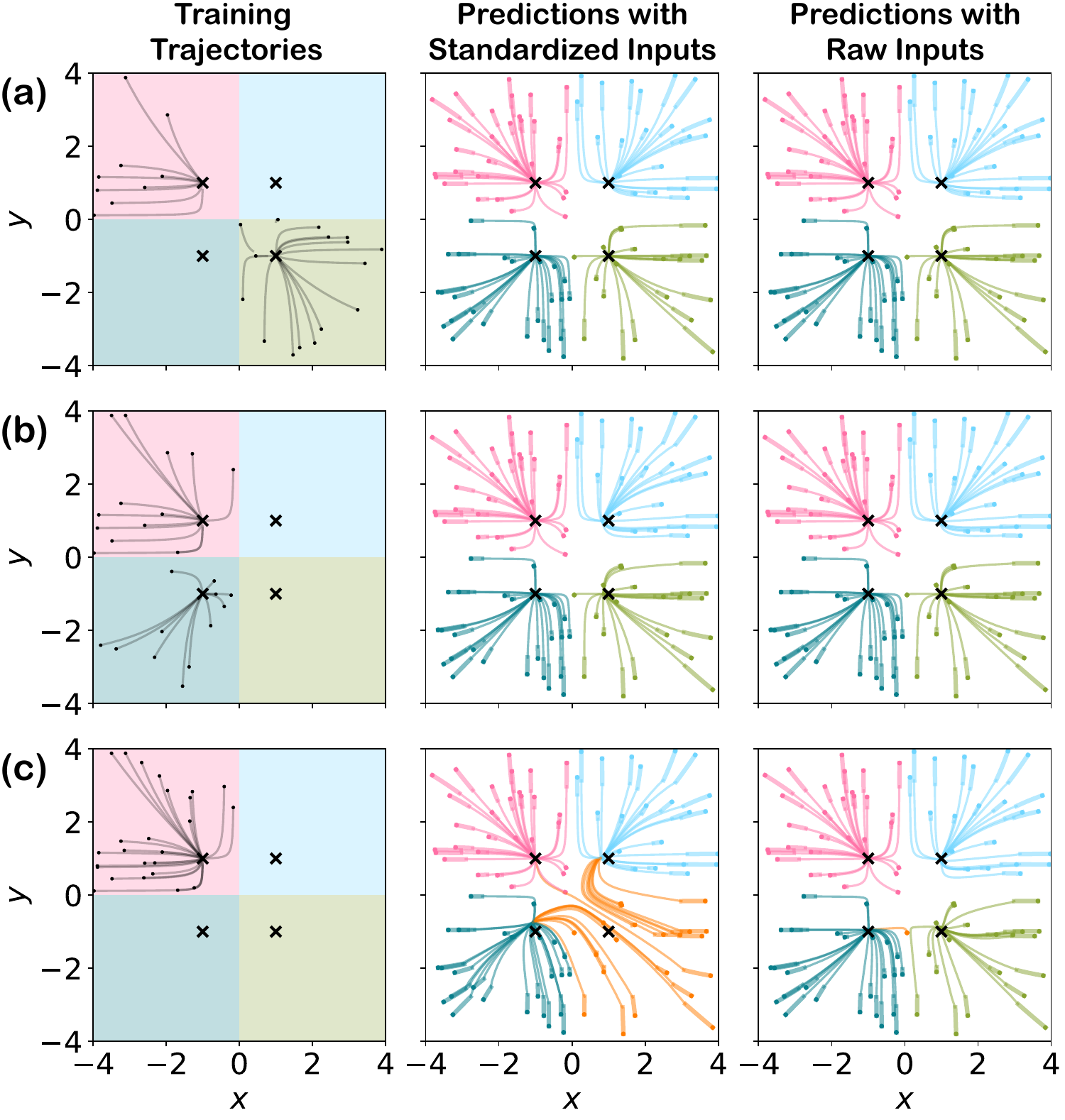}
    \caption{\justifying\textbf{RC identification of unseen attractors in a multi-well system with segregated basins.} We train an RC of ${N_r=200}$ nodes on ${N_{train}=25}$ fully-observed trajectories (gray lines) from two \textbf{(a, b)} or just one \textbf{(c)} of the multi-well system's basins of attraction \textbf{(left)}. Then, we forecast the system's evolution from $100$ initial conditions (dots) and corresponding short test signals (thick lines, ${N_{test}=5}$ observations) in two cases: when the inputs to RC are standardized \textbf{(center)}, and when the inputs to the RC are not standardized \textbf{(right)}. We color predictions (thin lines) orange if they go to an incorrect attractor, and according to their true basin otherwise.
    }
    \label{fig:Multi_Well_Phase_Space}
\end{figure}

Specifically, we employ a simple multi-well potential with four segregated basins of attraction. The dynamics are governed by two independent differential equations:
\begin{subequations}\label{eq:multi_well}
\begin{equation}\label{eq:wells_xdot}
    \dot{x}=\frac{1}{2}x(1-x^2),
\end{equation}
\begin{equation}\label{eq:wells_ydot}
    \dot{y}=\frac{1}{2}y(1-y^2),
\end{equation}
\end{subequations}
which yield four stable fixed-point attractors at the corners of the unit square and an unstable fixed point at the origin.

We demonstrate in \cref{fig:Multi_Well_Phase_Space} that our RC setup can generalize to unseen segregated basins in the multi-well system. However, the quality of this generalization is somewhat influenced by whether inputs to the reservoir are standardized based on the training distribution (a common practice in machine learning). Standardization schemes aim to ensure that each component of the inputs has a similar scale, which helps prevent a subset of inputs from dominating the dynamics and promotes more effective use of the system's computational capacity.  However, standardization schemes that are based on the training distribution implicitly bias the model to perform well on data with similar distributions (i.e., prioritize in-distribution performance). This bias may not hinder generalization to unseen intertwined basins of attraction, but we show in \cref{fig:Multi_Well_Phase_Space} that it can slightly degrade performance in unseen segregated basins of attraction, where the mean and range of the test data differ substantially from the training data.

In the central column of \cref{fig:Multi_Well_Phase_Space}, we use the input strength range $\sigma=1.0$ (\cref{tab:Reservoir_Parameters}), and we standardize all inputs such that the training inputs have component-wise mean zero and maximum absolute value one, as before. (It is also common to standardize inputs to have unit variance over the training data, but we use the maximum absolute value in this paper so that the standardization is not affected by long trajectories that are near a fixed point.) In the right-hand column, we do not standardize the training or test signal inputs and we scale the input strength range, $\sigma=0.25$, accordingly. When the range of the training trajectories approximately spans the range of the test initial conditions, ${-4\lesssim x\lesssim4}$ and ${-4\lesssim y\lesssim4}$ (row~a), the RC performs similarly with standardized and raw inputs. When the training trajectories sample only the left two basins of attraction, {$x<0$} (row~b), the RC again generalizes well in both cases, but captures the locations of the unseen fixed points at ${x=1}$ slightly more accurately when the input time series are not standardized. Finally, if the training data sample only one basin of attraction (row~c), the RC identifies all of the unseen attractors when the inputs are not standardized (raw) but fails to infer one of the unseen fixed-point attractors when the inputs to the reservoir are standardized.

%% file: sections/Magnetic_Pendulum_Results.tex
\subsection{Complex Basin Structure in a Magnetic Pendulum System}
\label{subsec:Magnetic}

\begin{figure*}[t]
	\centering
	\includegraphics[width=\linewidth]{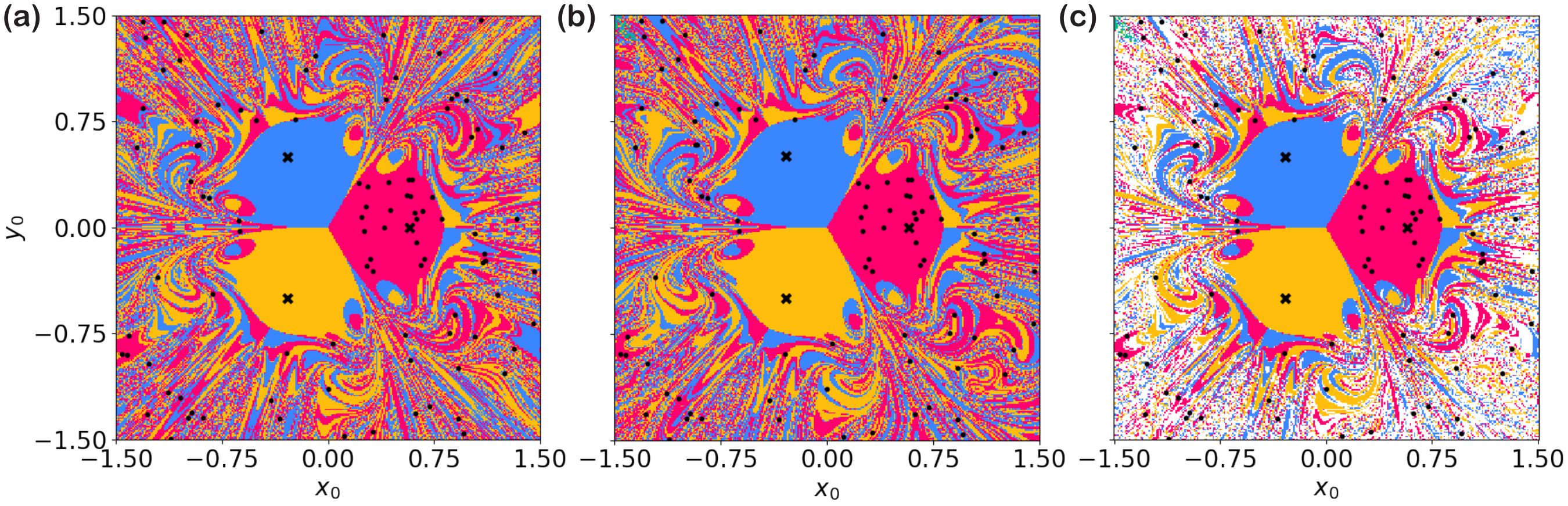}
	\caption{\justifying \textbf{An RC trained on data from only one basin can capture the magnetic pendulum's complex basin structure.} \textbf{(a)}~Basins of attraction of the magnetic pendulum in the plane ${\dot{x}=\dot{y}=0}$. Black crosses mark the positions of the magnets. \textbf{(b)}~RC-predicted basin structure using test signals of length ${N_{test}=100}$ data points with initial conditions on a ${300\times300}$ grid. The training and test signals are partially observed ($\dot{x}$ and $\dot{y}$ withheld); the RC learns to predict only the position components of the pendulum state. All $N_{train}=100$ training trajectories (black dots) are from the pink basin of attraction. Only ${f_{spurious}\approx8\times10^{-4}}$ of the predicted trajectories converge to spurious attractors (green). \textbf{(c)}~RC-predicted basins, with all initial conditions for which the RC predicts that the system converges to an incorrect fixed point (real or spurious) colored white.
    }
    \label{fig:Magnetic_Basins}
\end{figure*}

We now demonstrate that RCs can achieve similar out-of-domain generalization in a multistable system with a more complex basin structure than those of the Duffing or multi-well systems. The magnetic pendulum system\cite{Motter2013_DoublyTransient} consists of an iron bob suspended at the end of a pendulum above a plane that contains three magnetic point charges. The magnets sit at the vertices of an equilateral triangle and, when the pendulum hangs straight down, the bob is a height $d$ above the center of this triangle. We choose our coordinate system such that the origin is the triangle's center and the magnets' positions are ${(\Tilde{x}_1,\Tilde{y}_1)=(\frac{1}{\sqrt{3}}, 0)}$, ${(\Tilde{x}_2,\Tilde{y}_2)=(-\frac{1}{2\sqrt{3}}, -\frac{1}{2})}$, and  ${(\Tilde{x}_3,\Tilde{y}_3)=(-\frac{1}{2\sqrt{3}}, \frac{1}{2})}$. Taking the pendulum to be much longer than the distance between magnets (which is one), so that small angle approximations are applicable, the equations of motion are:
\begin{subequations}\label{eq:magnetic_pendulum}
\begin{equation}\label{eq:magnetic_xdot}
    \Ddot{x}=-\omega_0^2x-\gamma\dot{x}+\sum_{i=1}^3\frac{\Tilde{x}_i-x}{D_i(x,y)^3}
\end{equation}
\begin{equation}\label{eq:magnetic_ydot}
    \Ddot{y}=-\omega_0^2y-\gamma\dot{y}+\sum_{i=1}^3\frac{\Tilde{y}_i-y}{D_i(x,y)^3},
\end{equation}
\end{subequations}
where $\gamma$ is a damping coefficient, $\omega_0$ is the natural frequency of the pendulum, and
\begin{equation}
D_i(x,y)=\sqrt{(\Tilde{x}_i-x)^2+(\Tilde{y}_i-y)^2+d^2}
\label{eq:Magnet_Distances}
\end{equation}
is the distance from the bob to the $i^{\text{th}}$ magnet. The pendulum has three stable fixed points, each corresponding to it hanging directly above one of the three magnets. We choose the frequency ${\omega_0=0.5}$, damping ${\gamma=0.2}$, and pendulum height ${d=0.2}$, so that the system is dissipative and all trajectories, except for those on the stable manifold of an unstable fixed point at the origin (a set of measure zero), converge to one of the three stable fixed points.

The basin structure of the magnetic pendulum, which we plot in the plane ${\dot{x}=\dot{y}=0}$ in \cref{fig:Magnetic_Basins}a, is considerably more complex than that of the Duffing system. In fact, while not a true fractal, the basin boundary forms a fractal-like structure\cite{Motter2013_DoublyTransient} -- a so-called `slim fractal.'\cite{Chen2017_SlimFractals} The resulting sensitivity of the pendulum's final state to small perturbations makes basin prediction considerably more difficult. 
Despite the challenge posed by transient chaos, however, we demonstrate in \cref{fig:Magnetic_Basins}b that an RC trained on trajectories from only one of the magnetic pendulum's basins of attraction can nonetheless generalize to the other unexplored basins.

The complexity of the pendulum's basin structure means that small forecast errors can easily push the closed-loop RC system into an incorrect basin of attraction. As a result, we find that to achieve good performance, compared with the Duffing and multi-well systems, our setup for the magnetic pendulum requires: (1) more training trajectories and a more powerful (i.e., larger) reservoir to improve the output-layer accuracy and (2) longer test signals to improve the RC's initial synchronization. In \cref{fig:Magnetic_Basins}, for example, we train a reservoir of size ${N_r=2500}$ nodes on ${N_{train}=100}$ trajectories, and make predictions using test signals of length ${N_{test}=100}$ data points. (We demonstrate in \cref{fig:Magnetic_Basins_TestLengths} of the supplementary material that these test signals are still short enough such that predicting the correct basin from the end of the test signal is nearly as difficult as predicting it from the initial condition (start of the test signal). We also show how the RC's performance varies with the length of the test signals.) Remarkably, despite the fractal-like basin boundaries, the RC is able to provide good predictions not only for the training (pink) basin but also for the two other unseen basins. Still, because of the difficulty of the problem, we do not expect that any RC -- even one trained on trajectories from all basins -- can reliably predict the correct attractor for initial conditions far from the fixed points. (Even a next-generation reservoir computer with a very strong structural prior requires training data with a high sampling rate to reliably predict the basins of the magnetic pendulum.\cite{Zhang2023_Catch22}) \cref{fig:Magnetic_Basins} illustrates this inherent challenge. The RC-predicted basin structure (\cref{fig:Magnetic_Basins}b) matches the true basin structure (\cref{fig:Magnetic_Basins}a) well qualitatively, even as the RC struggles to reliably predict the basins of individual test signals whose initial conditions are in the outer regions of the test grid, where the basin structure is most complex (\cref{fig:Magnetic_Basins}c). The fraction of test signals for which the RC predicts the correct attractor, $f_c\approx0.66$, however, is still substantially higher than the baseline fraction of $0.47$ achieved by an approach that simply guesses that the pendulum will converge to the magnet nearest to it at the end of the test signal (supplementary material \cref{fig:Magnetic_Basins_TestLengths}).

Assessing the qualitative accuracy of the predicted basins in \cref{fig:Magnetic_Basins}b is analogous to evaluating traditional climate replication in monostable dynamical systems, which focuses on capturing statistical properties of the system over time. Here, our goal is for the statistics of the predicted system behavior -- collected over different test signals -- to accurately reflect the system's multistability. Indeed, we see that the RC-predicted trajectories rarely converge to spurious attractors ($f_{spurious}\approx8\times10^{-4}$, i.e., $75$ of the $90000$ sample predictions in \cref{fig:Magnetic_Basins}) and the RC broadly captures where extended regions of state space belong to the same basin of attraction and where the basins are more intertwined.

\begin{table}[b]
\centering
\caption{\justifying
Per-basin RC performance in predicting the correct attractor in \cref{fig:Magnetic_Basins}.  
Baseline: a predictor that always guesses the magnet nearest the pendulum at the end of the test signal.}
\label{tab:Magnetic_Basin_Performance}
\resizebox{\columnwidth}{!}{%
\begin{ruledtabular}
\begin{tabular}{l|ccc}
Basin & Fraction Correct & False Neg. Rate & False Pos. Rate \\
\hline
Pink   & 0.66 & 0.34 & 0.17 \\
Blue   & 0.65 & 0.35 & 0.17 \\
Yellow & 0.66 & 0.34 & 0.18 \\
\hline
Baseline & 0.47 & -- & -- \\
\end{tabular}
\end{ruledtabular}
}
\end{table}

\begin{figure*}[!th]
    \centering
    \includegraphics[width=\linewidth]{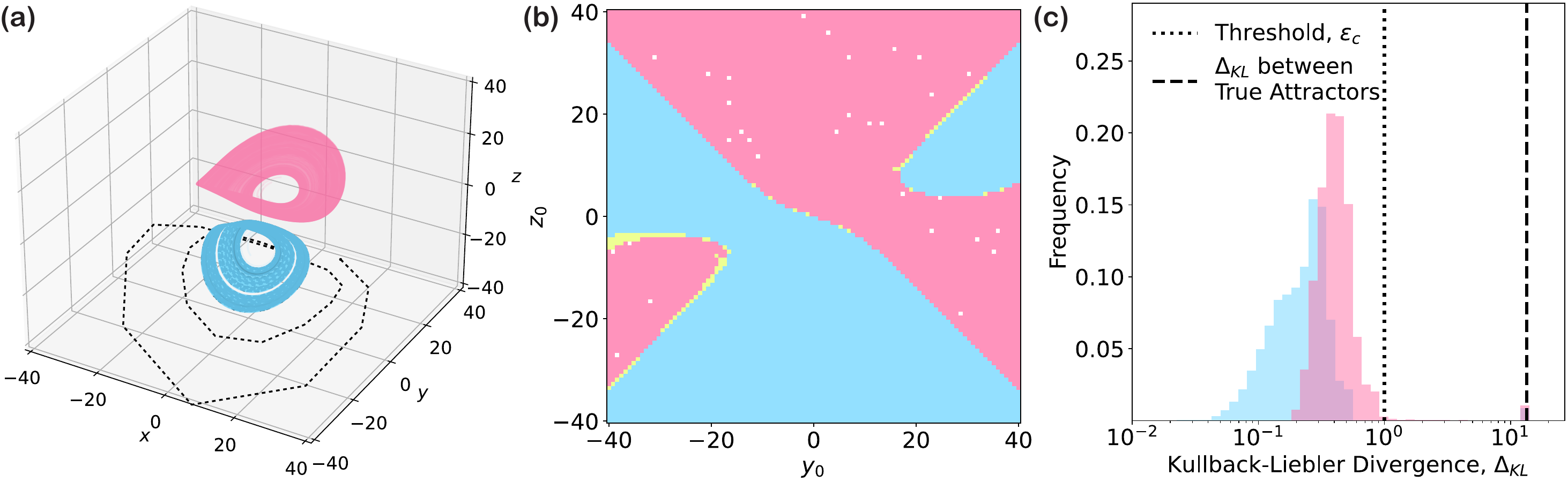}
    \caption{\justifying\textbf{RC reconstruction of an unseen chaotic attractor in a multistable Lorenz-like system.} We train an RC with ${N_r=500}$ nodes on a single fully-observed trajectory (dashed line) of the Lorenz-like system. \textbf{(a)} We forecast the system's evolution from $36$ short test signals (${N_{test}=50}$ observations, test signal trajectories not shown for clarity), each starting from a random initial condition. Forecasts start from the endpoints of these test signals, and each predicted trajectory is colored according to the true basin of its initial condition. \textbf{(b)} We make predictions from shorter test signals (${N_{test}=5}$) with initial conditions forming a ${100\times100}$ grid in the $x=0$ plane, and plot the RC-predicted basin structure. Yellow and white, respectively, indicate initial conditions for which the prediction converges to the incorrect attractor and those for which the prediction does not accurately resemble either true attractor (${\Delta_{KL}(P_{A_i},P_{\hat{\boldsymbol{u}}})>\epsilon_c}$ for ${i=1,2}$). \textbf{(c)} Distribution of the KL divergence between the predicted and true attractors, computed over all predictions in panel (b). Each histogram is formed from predictions associated with test signals from the same true basin of attraction and is colored accordingly. The vertical dotted line indicates the convergence threshold, ${\varepsilon_c=1}$, and the vertical dashed line marks the KL divergence between the true attractors, estimated as ${\frac{1}{2}\left(\Delta_{KL}(P_{A_1},P_{A_2})+\Delta_{KL}(P_{A_2},P_{A_1})\right)}$.
    }
    \label{fig:MLorenz_Results}
\end{figure*}

At the per-basin level, the fraction correct ($f_c$), false negative rate ($1-f_c$), and false positive rate are summarized in \cref{tab:Magnetic_Basin_Performance}. These results, coupled with \cref{fig:Magnetic_Heatmaps} of the supplementary material, indicate that RC performance in predicting basin membership is comparable for seen and unseen basins, and that, in many cases, an RC trained on trajectories from only one basin of the magnetic pendulum still recovers the overall basin structure almost as well as an RC trained on all three basins -- consistent with what we find for the Duffing system (\cref{fig:Duffing_Heatmaps}). However, even when the RC is successful at predicting basin membership, it may still miss some of the detailed dynamics of the unseen basins (see supplementary material \cref{fig:Magnetic_Error_vs_Time}). Overall, these findings suggest that an RC can learn the system dynamics in a way that enables generalization to unseen regions of state space in an operationally useful manner, even without strictly capturing system symmetries.

%% file: sections/Lorenz_Results.tex
\subsection{Coexisting Chaotic Attractors in a  Lorenz-like System}
\label{subsec:Lorenz}

The Duffing, multi-well, and magnetic pendulum systems are dissipative, with all trajectories converging to fixed-point attractors. In many multistable systems, however, the system state never reaches a fixed point and instead converges to a limit cycle or chaotic attractor, along which it evolves indefinitely. In this section, we highlight a case in which an RC captures the existence of an unseen chaotic attractor.

Following L\"u et al.,\cite{Lu2004_MultistableLorenz} we study a multi-stable Lorenz-like system that is governed by three coupled ordinary differential equations:
\begin{subequations}\label{eq:Lorenz}
\begin{equation}\label{eq:Lorenz_x1}
\dot{x}=-\frac{ab}{a+b}x-yz+c,
\end{equation}
\begin{equation}\label{eq:Lorenz_x2}
\dot{y}=ay+xz,
\end{equation}
\begin{equation}\label{eq:Lorenz_x3}
\dot{z}=bz+xy.
\end{equation}
\end{subequations}
When ${a=-10}$, ${b=-4}$, and ${c=18.1}$, the system has two chaotic attractors. In \cref{fig:MLorenz_Results}a, we train an RC on a single long trajectory (black dashed line), which approaches and then evolves along one of the attractors. Then we show that the RC captures the existence and structures of both the seen attractor (the blue lobe, ${A_1}$) and the unseen attractor (the pink lobe, ${A_2}$). To obtain a clear image of the predicted chaotic attractors in \cref{fig:MLorenz_Results}a, we make predictions using test signals of ${N_{test}=50}$ data points and omit the test signals themselves from the figure. These test signals are, however, long enough that each prediction starts when the system state has already reached one of the chaotic attractors.

In \cref{fig:MLorenz_Results}b, we reduce the length of the test signals to ${N_{test}=5}$ observations so that, for most initial conditions, the system is still far from the attractors when prediction begins. Then we plot the RC-predicted basin structure for test signals with initial conditions in the plane $x=0$. The RC accurately captures the basin structure of the multistable Lorenz-like system. In fact, the RC predicts the correct attractor for almost all initial conditions in the ${100\times100}$ test grid, indicating that it can accurately capture the system's transient behavior en route to the chaotic attractors (in addition to reconstructing the chaotic attractors themselves).

The convergence criteria used for the Duffing, multi-well, and magnetic pendulum systems in \cref{subsec:Duffing,subsec:Multi_well,subsec:Magnetic} do not apply to the Lorenz-like system because its attractors are not fixed points. Instead, to determine whether a trajectory ${\boldsymbol{u}(t)}$ converges to one of the system's chaotic attractors, we compute the Kullback-Liebler (KL) divergence of the empirical state distribution, $P_{\boldsymbol{u}}(\boldsymbol{v})$, over the final 500 time steps of ${\boldsymbol{u}(t)}$ relative to the distribution $P_A(\boldsymbol{v})$ from a reference trajectory on the attractor:
\begin{equation}\Delta_{KL}\left(P_A,P_{\boldsymbol{u}}\right)=\int_{\boldsymbol{v}\,\in\,\mathbb{R}^3}P_A(\boldsymbol{v})\log\frac{P_A(\boldsymbol{v)}}{P_{\boldsymbol{u}}(\boldsymbol{v})}d\boldsymbol{v}.
\label{eq:KL_Divergence}
\end{equation}
Following previous work,\cite{Zhang2025_Zeroshot,Hess2023_TeachingForcing} we use Gaussian mixture models to estimate the distributions ${P_{\boldsymbol{u}}(\boldsymbol{v})}$ and ${P_A(\boldsymbol{v})}$ from the predicted and reference trajectories. If the KL divergence of the predicted distribution relative to the true distribution on the attractor $A$ is less than a threshold value, ${\Delta_{KL}(P_A,P_{\boldsymbol{u}})<\varepsilon_c}$, we say that the predicted trajectory converges to the attractor $A$.

In \cref{fig:MLorenz_Results}c, we use the KL divergence to assess the quality of the reconstructed attractors in the seen and unseen basins. The RC achieves comparable accuracy on the unseen (pink) attractor as on the seen (blue) attractor, with the KL divergence of the predicted attractor relative to the true attractor in both cases substantially lower than the KL divergence between the two true attractors (vertical dashed line).

%% file: sections/Discussion.tex
\section{Discussion}\label{sec:Discussion}

Our results show that reservoir computers (RCs) can successfully generalize to entirely unobserved regions of state space in multistable dynamical systems. Unlike approaches that rely on explicit system knowledge or dense coverage of the training domain, our RC setup requires only a limited set of observed trajectories and makes no assumptions about the underlying dynamics -- i.e., it operates without explicit structural priors, such as known equations, symmetries, or conservation laws -- yet still learns representations that support strong out-of-domain generalization.

We make use of a training scheme that allows the RC to incorporate information from multiple disjoint time series. Importantly, we show that RCs trained on trajectories from a single basin of attraction can accurately predict system behavior in other basins. After training, the RC can generate predictions from a new initial condition and observed signal -- one that only needs to be long enough for the reservoir to synchronize -- without needing to retrain or reconfigure the model.

A strength of this approach is that it enables generalization from a relatively limited set of training trajectories -- limited both in number and in the portion of state space sampled -- notably more restricted than typically used in data-intensive machine learning frameworks. However, successful generalization still depends on whether the training data contain sufficient information about the system’s dynamics. If the training trajectories fail to capture a sufficient range of the system's behaviors, the RC will not have enough information to represent behavior beyond the training domain, and generalization will break down.

These findings suggest that RCs can construct a generalizable internal representation of system dynamics that extends beyond the training data. In doing so, they can capture global structures such as basins of attraction and make reliable forecasts in regions of state space that were never directly observed. This kind of generalization -- achieved from sparse, disjoint training data and without structural assumptions -- positions RCs as a practical tool for modeling complex systems in settings where prior knowledge is unavailable and data are limited.

Moving forward, a promising direction for future work is to develop a mathematical understanding on what enables out-of-domain generalization in RCs. Generally speaking, vector fields in part of the state space do not uniquely determine vector fields in other parts of the state space, so some kind of inductive bias is needed to accurately predict dynamics far away from the training data. Unlike most neural networks trained via gradient descent, RC output weights are determined through regularized linear regression -- a convex optimization problem with a closed-form solution. In the overparameterized regime, the Moore–Penrose pseudoinverse in the closed-form solution selects the solution with the smallest norm. This optimization process may introduce an implicit inductive bias that favors simpler, lower-complexity solutions. In many systems,  such simplicity may be an effective inductive bias that enables generalization beyond the training domain. Exploring how different regularization schemes influence this tendency and how they connect to broader ideas in machine learning -- such as flat minima\cite{Feng2021_FlatMinima} and double descent\cite{Belkin2019_DoubleDescent,Ribeiro2021_DoubleDescentModelingDynamics} -- offers a promising direction for deepening our understanding of generalization in RCs.

%% file: sections/Supplemental_Statement.tex
\section*{Supplementary Material}

See the supplementary material for \cref{fig:Full_Duffing_Trajectories,fig:Duffing_Dynamical_Noise,fig:Magnetic_Basins_TestLengths,fig:Magnetic_Error_vs_Time,fig:Magnetic_Heatmaps}, which provide further information on the sensitivity of RC generalization to dynamical/process noise, test signal length, and training data diversity, as well as illustrations of predicted near-attractor behavior and spurious attractors.

%% file: sections/Acknowledgements.tex
\begin{acknowledgments}\label{sec:Acknowledgements}
We thank Edward Ott and Brian Hunt for helpful conversations, insights, and suggestions. We also acknowledge the University of Maryland supercomputing resources (\href{http://hpcc.umd.edu}{http://hpcc.umd.edu}) made available for conducting the research reported in this paper. 
The contributions of D.N. and of M.G. were supported, respectively, by the National Science Foundation Graduate Research Fellowship Program under Grant No.\ DGE 1840340 and by ONR Grant No.\ N000142212656. 
Y.Z. was supported by the Omidyar Fellowship and the National Science Foundation under Grant No.\ DMS 2436231.
Any opinions, findings, and conclusions or recommendations expressed in this material are those of the authors and do not necessarily reflect the views of the National Science Foundation, the Department of Defense, or the U.S. Government.
\end{acknowledgments}

%% file: sections/Author_Declarations.tex
\section*{Author Declarations}

\subsection*{Conflict of Interest}

The authors have no conflicts to disclose.

\subsection*{Author Contributions}

\noindent\textbf{D.~A.~Norton:}~Conceptualization (equal); Formal analysis (lead); Funding acquisition (equal);  Investigation (lead); Methodology (equal); Software (lead); Validation (lead); Visualization (lead); Writing -- original draft (lead). \textbf{Y.~Zhang:}~Conceptualization (equal); Funding acquisition (equal); Methodology (equal); Visualization (supporting); Writing -- review \& editing (equal). \textbf{M.~Girvan:}~Conceptualization (equal); Funding acquisition (equal); Methodology (equal); Project administration (lead); Resources (lead); Supervision (lead); Visualization (supporting); Writing -- review \& editing (equal).

\subsection*{Data Availability}

%The data that support the findings of this study are available from the corresponding author upon reasonable request.

The code that supports the findings of this study is available at the following repository:\\ \\\href{https://github.com/nortondeclan/Learning_Beyond_Experience}{https://github.com/nortondeclan/Learning\_Beyond\_Experience}.

%% file: sections/Bibliography.tex
\section{References}
\bibliography{LARC_Bibliography.bib}

%% file: sections/Appendix.tex
\appendix*
\section{Experimental Setup}
\label{sec:Experimental_Setup}

We obtain trajectories of the Duffing, multi-well, and multistable Lorenz-like systems by integrating \cref{eq:Duffing,eq:multi_well,eq:Lorenz} using a fourth order Runge-Kutta integrator. We generate trajectories of the magnetic pendulum system by integrating \cref{eq:magnetic_pendulum} using using the \textsc{scipy} integrate.solve\_ivp implementation of the DOP853 eigth order Runge-Kutta integration scheme. For the magnetic pendulum, the integration time step is adaptive and for the other systems, it is fixed (${\Delta t=0.01}$). In all cases, the time step between samples in the RCs' training and test signals is fixed: ${\Delta t=0.01}$ for the Duffing and multi-well systems and ${\Delta t=0.02}$ for the magnetic pendulum and multistable Lorenz-like systems, as in \cref{tab:Experimental_Parameters}. Because we intend for the RC to learn from and predict the transient dynamics of each system, we do not discard any portion of the integrated trajectories before forming training or test signals. (We do however, discard a short transient response of the reservoir's internal state at the start of each training signal, as described in \cref{sec:Methods}.)

\begin{table}[h]
\caption{\justifying Experimental Parameters.}
\begin{ruledtabular}
\begin{tabular}{ll|cccc}
& & Duffing & Wells & Magnetic & Lorenz \\\hline
RC Time Step & $\Delta t$ & $0.01$ & $0.01$ & $0.02$ & $0.02$ \\
Training Transient & $N_{trans}$ & 5 & 5 & 25  & 5\\
Training Signal Length & $N_i\ \forall\ i$ & 500 & 500 & 500 & 5000\\
Forecast Horizon & $N_f$ & 2000 & 2000 & 2000 & 5000\\
Convergence Threshold & $\varepsilon_c$ & 0.5 & -- & 0.25 & 1.0
\end{tabular}
\end{ruledtabular}
\label{tab:Experimental_Parameters}
\end{table}

The parameters we use to construct and train RCs for our experiments are provided in \cref{tab:Reservoir_Parameters}. The other parameters defining our experiments are in \cref{tab:Experimental_Parameters}. For simplicity, we use training trajectories that are of equal length, ${N_i=500\ \forall\ i=1,...,N_{train}}$, in all of our experiments (except for the multistable Lorenz-like system, which we address separately in the next paragraph). Our multiple-trajectory training scheme (\cref{sec:Methods}) does not, however, require that all training signals are the same length. When all training trajectories must be from the same basin of attraction, we first generate a trajectory of length $4000\Delta t$ from each sample initial condition and check whether the trajectory converges to the corresponding desired attractor. If a trajectory converges to the right attractor, we include its first ${N_i=500}$ data points in the RC's training data. We repeat this process with new sample initial conditions until we obtain the desired number of training trajectories, $N_{train}$, from the chosen basin. In all of our experiments (except the Lorenz-like system), we forecast from the end of each provided test signal, i.e., ${t=(N_{test}-1)\Delta t}$, to time ${t=N_f\Delta t=2000\Delta t}$.

\begin{table}%[t]
\caption{\justifying Reservoir Hyperparameters.}
\begin{ruledtabular}
\begin{tabular}{ll|cccc}
& & Duffing & Wells & Magnetic & Lorenz \\\hline
Reservoir Size & $N_r$ & 200 & 200 & 2500 & 500\\
Mean In-degree & $\langle d\rangle$ & 10 & 10 & 10 & 10 \\
Input Strength Range & $\sigma$ & 1.0 & 1.0 \footnote{We use ${\sigma=1.0}$ when we standardize inputs to the reservoir and ${\sigma=0.25}$ when we do not standardize the inputs.} & 5.0 & 0.5\\
Spectral Radius & $\rho$ & 0.4 & 0.4 & 0.4 & 0.4\\
Bias Strength Range & $\psi$ & 0.5 & 0.5 & 0.5 & 0.5\\
Leakage Rate & $\lambda$ & 1.0 & 1.0 & 1.0 & 1.0\\
Tikhonov Regularization & $\alpha$ & $10^{-12}$ & $10^{-12}$ & $10^{-10}$ & $10^{-10}$\\
Training Noise Amplitude & $\eta$ & $10^{-5}$ & $10^{-5}$ & $10^{-3}$ & $10^{-3}$
\end{tabular}
\end{ruledtabular}
\label{tab:Reservoir_Parameters}
\end{table}

For our experiments with the multistable Lorenz-like system, we use only one training trajectory, which is of length ${N_1=5000}$ to allow for adequate sampling of the on-attractor chaotic dynamics. Similarly, for each test signal, we forecast from ${t=(N_{test}-1)\Delta t}$, to time ${t=N_f\Delta t=5000\Delta t}$. To assess whether a prediction converges to one of the system's true attractors, we measure the Kullback-Liebler (KL) divergence of the empirical distribution of Lorenz states over the final $500$ time steps of the predicted trajectory relative to that of a reference trajectory on the attractor, as described in \cref{subsec:Lorenz}. We measure the KL divergences using only the final $500$ time steps of the predicted trajectory to ensure that (1) the transient approach to the attractor does not obscure the on-attractor behavior and (2) the RC's prediction is stable and has remained on the attractor for a prolonged period (the previous ${4500-N_{test}}$ predicted data points). To estimate the KL divergence, we use the \textsc{dysts}\cite{Gilpin2021_ChaosBenchmark,Gilpin2023_ModelScaleDomainKnowledge} metrics.estimate\_kl\_divergence routine with its default arguments: ${\text{n\_samples}=1000}$, ${\text{sigma\_scale}=1}$, and ${\text{eps}=10^{-10}}$.

As shown in \cref{tab:Reservoir_Parameters}, the reservoir hyperparameters that we use for our experiments with the Duffing, multi-well, magnetic pendulum, and Lorenz-like systems are identical except for the input strength range, $\sigma$, regularization strength, $\alpha$, and noise amplitude, $\eta$. We chose these hyperparameters by coarse hand-tuning to allow for good, but not necessarily optimal performance. We chose the other hyperparameters to have values that typically allow for reasonably accurate forecasting with reservoir computers, and performed no experiment-specific tuning of these values. While more robust hyperparameter tuning may improve performance overall, our priority is to demonstrate that reservoir computers can generalize to unexplored regions of state space without system-specific structural constraints, rather than to obtain highly optimized forecasts.

%% file: sections/Supplemental.tex
\setcounter{figure}{0} % Reset figure counter
\let\oldthefigure\thefigure % Capture figure numbering scheme
\renewcommand{\thefigure}{S\oldthefigure}
\newpage

\begin{figure*}[!th]
    \centering
    \includegraphics[width=\linewidth]{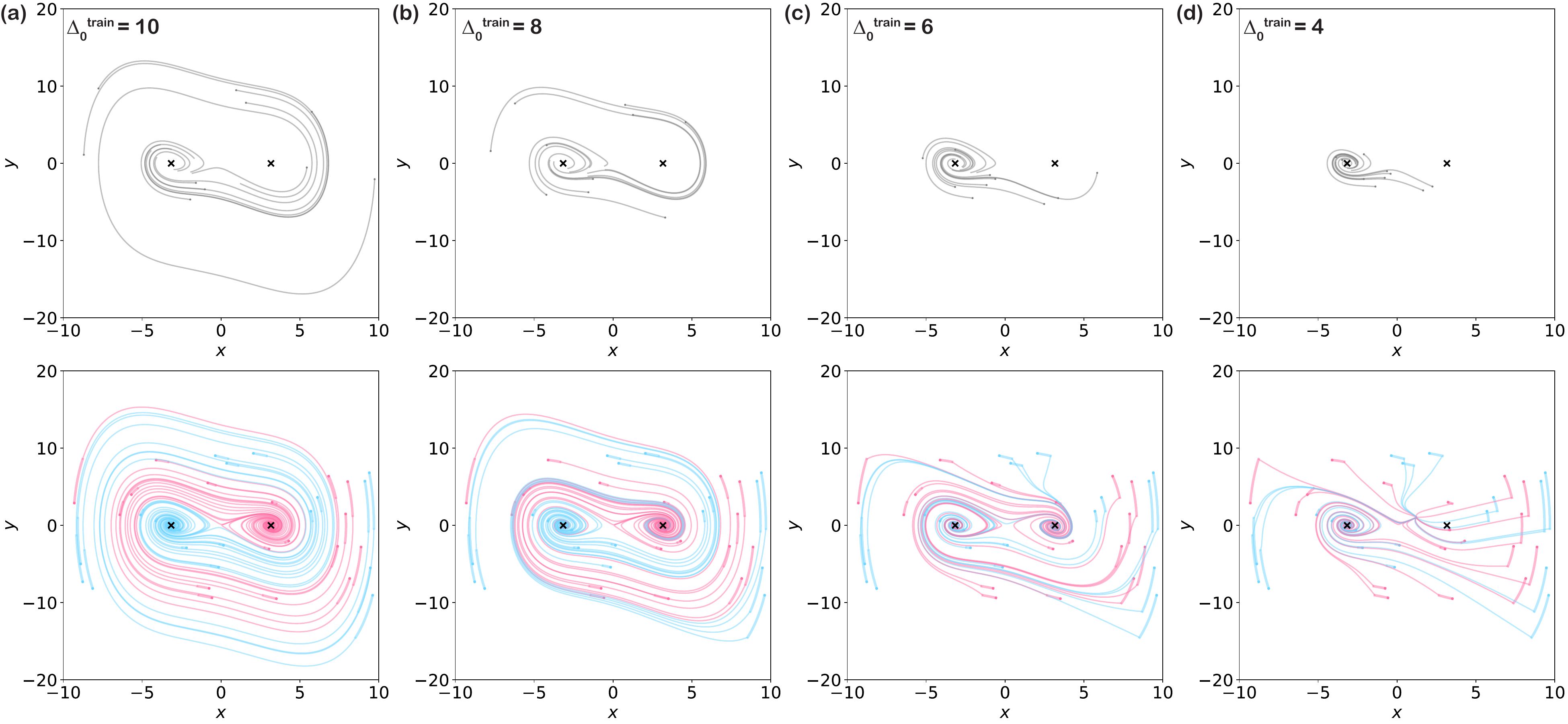}
    \caption{\justifying \textbf{RCs can reliably generalize to an unseen basin of the Duffing system, unless the training data are very restricted.} We train a reservoir computer (RC) of ${N_r=200}$ nodes to forecast fully-observed states of the Duffing system. We draw ${N_{train}=10}$ training signals (top row) from the basin of attraction ${B(\boldsymbol{A}^-)}$ for the fixed point ${\boldsymbol{A}^-}$, and make predictions from $36$ test signals of length ${N_{test}=10}$ points, whose initial conditions are drawn randomly from ${-10\leq x_0\leq10}$ and ${-10\leq y_0\leq10}$ and from both basins of attraction, ${B(\boldsymbol{A}^\pm)}$. We restrict the training initial conditions to the ranges ${-\Delta_0^{train}\leq x_0\leq\Delta_0^{train}}$ and ${-\Delta_0^{train}\leq y_0\leq\Delta_0^{train}}$. In \textbf{(a)} to \textbf{(d)} we decrease the half-width of the range of the training initial conditions, from ${\Delta_0^{train}=10}$ (a) to ${\Delta_0^{train}=4}$ (d), while holding the half-width of the range of the test initial conditions, ${\Delta_0^{test}=10}$, fixed. As ${\Delta_0^{train}}$ decreases, the RC's performance away from the fixed points diminishes. When ${\Delta_0^{train}=4}$, the RC predicts the existence of a spurious attractor, which does not coincide with the position of the true unseen attractor, ${\boldsymbol{A}^+\approx(3.16,0)}$.}
    \label{fig:Full_Duffing_Trajectories}
\end{figure*}

\begin{figure*}[!th]
    \centering
    \includegraphics[width=\linewidth]{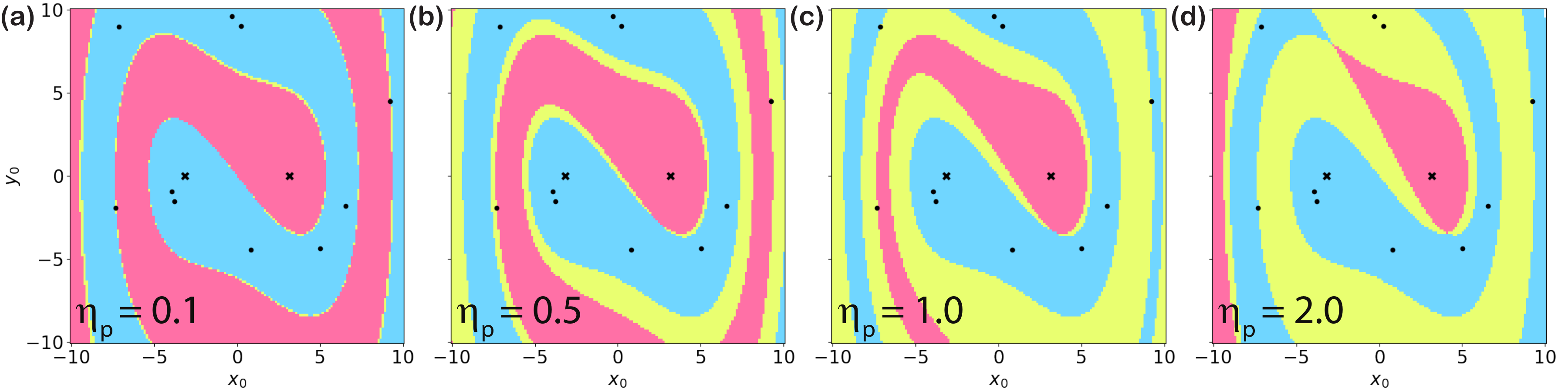}
    \caption{\justifying\textbf{RCs can reliably generalize to an unseen basin of the Duffing system, even when moderate stochastic forcing (dynamical/process noise) influences the training trajectories.} We train an RC of ${N_r=200}$ nodes on ${N_{train}=10}$ partially-observed trajectories ($x$ only) of the unforced Duffing system from the basin of attraction $B(\boldsymbol{A}^-)$ (blue). We integrate dynamical/process noise into the training trajectories by including white noise in the Duffing system's equations of motion: ${\dot{x}=y+\mathcal{N}(0,\eta_p)}$, ${\dot{y}=ay-bx-cx^3+\mathcal{N}(0,\eta_p)}$, where ${\mathcal{N}(0,\eta_p)}$ draws a random variable from a Gaussian distribution with mean zero and standard deviation $\eta_p$. The noise is independent for each component of the system and at each integration time step. We then make predictions from short noiseless test signals (${N_{test}=10}$ observations each) from both basins of attraction. \textbf{(a) to (d)}: The RC-predicted basins for different values of $\eta_p$, from ${\eta_p=0.1}$ (a) to ${\eta_p=2.0}$ (d). The blue/pink initial conditions are those that the RC correctly predicts belong to the seen/unseen basin. The initial conditions of test signals for which the RC-predicted trajectory converges to the incorrect attractor are yellow. Crosses mark the true fixed point attractors and dots mark the training initial conditions. We use $y_0$ for plotting purposes only; the RC has access to and then predicts only the $x$-component of the Duffing system.}
    \label{fig:Duffing_Dynamical_Noise}
\end{figure*}

\begin{figure*}[!t]
    \centering
    \includegraphics[width=.65\linewidth]{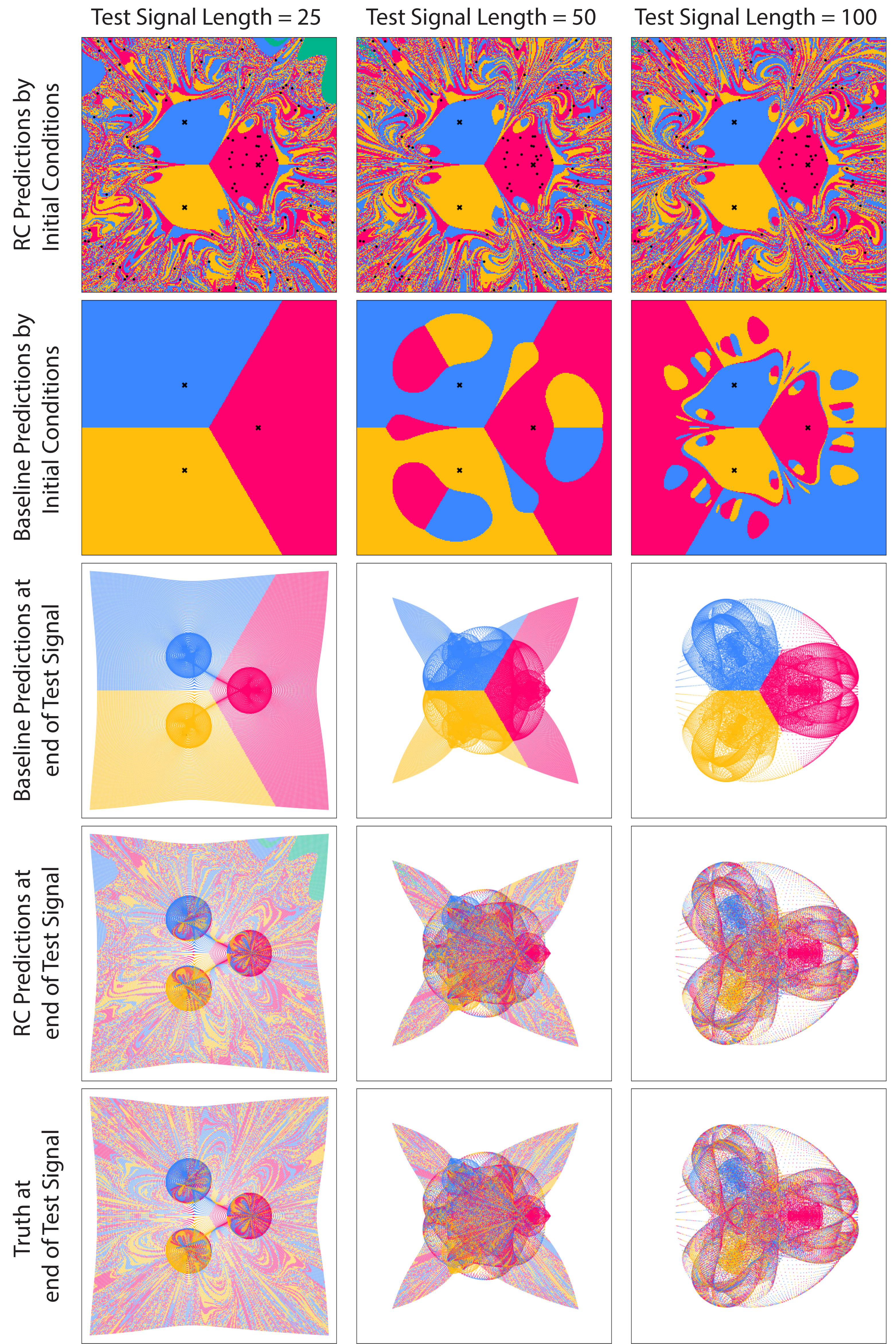}
    \caption{\justifying \textbf{RC Basin Predictions and Comparison with Baseline Across Test Signal Lengths.} \textbf{\textit{Row~1:}} Reservoir computer (RC) basin predictions over a ${300\times300}$ grid of initial conditions, i.e., the state at the start of the test signal. Each point is colored by the attractor to which the RC predicts convergence (pink, blue, yellow). Green points indicate that the RC predicts convergence to a spurious attractor. Fractions correctly classified: ${f_c\approx0.52}$ (${N_{test}=25}$), ${f_c\approx0.57}$ (${N_{test}=50}$), ${f_c\approx0.66}$ (${N_{test}=100}$). \textbf{\textit{Row~2:}} Baseline predictions over the same grid of initial conditions, where each point is assigned to the nearest fixed point based on its state at the end of the test signal (i.e., the start of the prediction interval). Fractions correct: ${f_c\approx0.45}$ (${N_{test}=1}$, not plotted), ${f_c\approx0.45}$ (${N_{test}=25}$),  ${f_c\approx0.45}$ (${N_{test}=50}$), ${f_c\approx0.47}$ (${N_{test}=100}$). \textbf{\textit{Rows~1~and~2}} show full-color basin maps spanning the entire initial condition grid. \textbf{\textit{Rows~3–5:}} Predictions and ground truth at the locations of the measured state at the end of each test signal. These panels contain only discrete points (colored dots) at those final states, with white backgrounds elsewhere. \textbf{\textit{Row~3:}} Baseline predictions for the final test signal states. \textbf{\textit{Row~4:}} RC predictions for the final test signal states. \textbf{\textit{Row~5:}} Ground-truth basin assignments at the final states of the test signals. When visualizing and comparing the RC predictions (row 4) vs ground truth (row 5) at the final states of the test signals, we see a similar trend as we saw when visualizing the basin maps from the locations of the initial conditions (\cref{fig:Magnetic_Basins} and row 1): the RC predictions and the ground truth have qualitatively similar basin structures. These plots, along with the calculated fractions correct, also indicate that the baseline method -- assigning each point to the nearest basin at the end of the test signal -- shows only modest improvement across the test signal lengths considered.
    }
    \label{fig:Magnetic_Basins_TestLengths}
\end{figure*}

\begin{figure*}[!t]
    \centering
    \includegraphics[width=.6\linewidth]{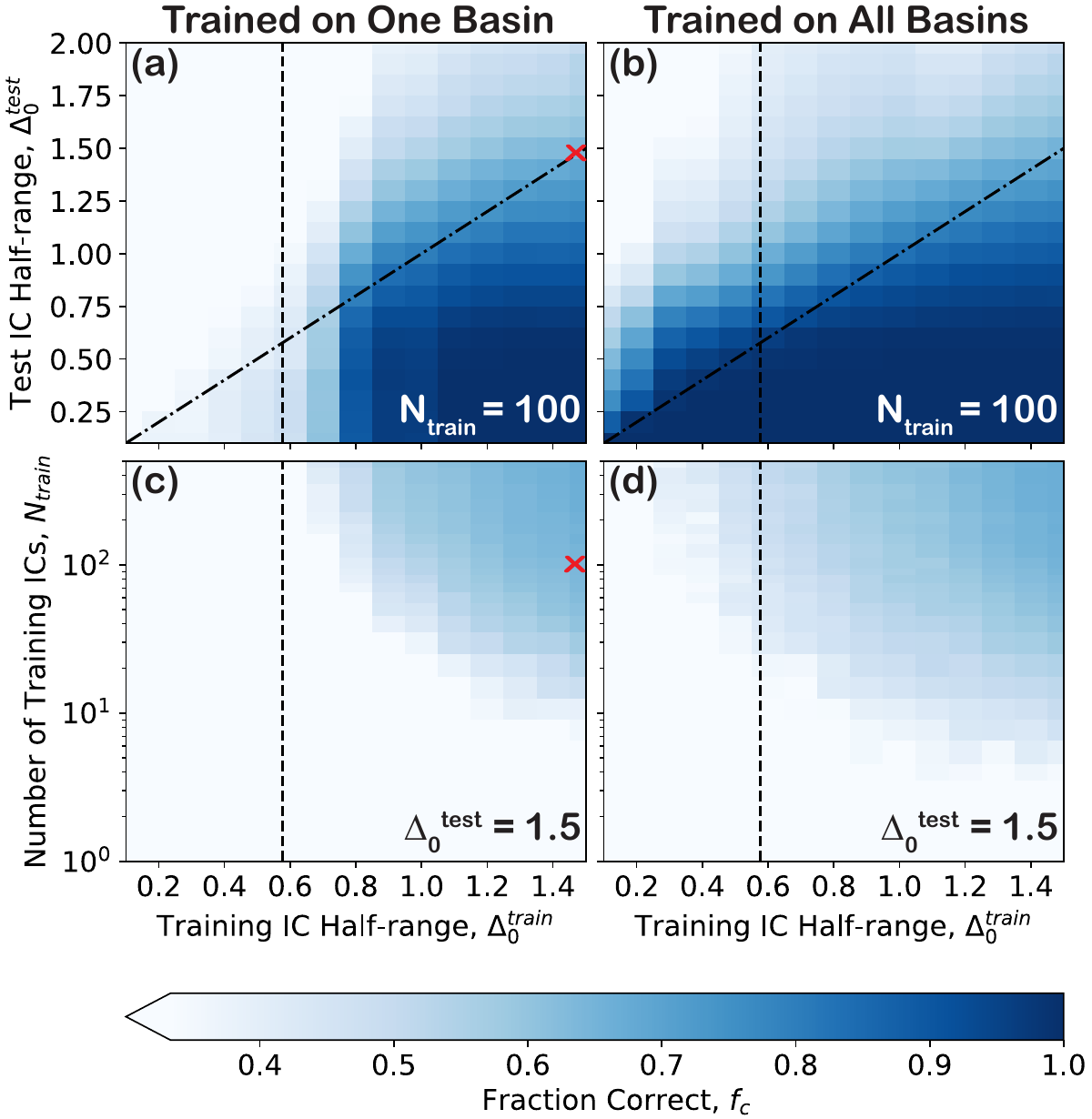}
    \caption{\justifying \textbf{In many scenarios, an RC generalizing to unseen basins of attraction of the magnetic pendulum performs comparably to one trained on data from all three of its basins.} We train an RC of ${N_r=2500}$ nodes to forecast partially-observed states of the magnetic pendulum system (the position components $x$ and $y$ only) and plot the fraction of initial conditions for which the RC predicts the correct basin of attraction as we vary the number of training initial conditions, $N_{train}$, and the half-widths of the ranges of the training and test initial conditions, ${\Delta_0^{train}}$ and ${\Delta_0^{test}}$. In \textbf{(a) and (c)}, we draw all training signals from the basin of attraction of the magnet at ${(\tilde{x}_1,\tilde{y}_1)=(1/\sqrt{3},0)}$, and in \textbf{(c) and (d)}, we draw training signals from all basins of attraction. In all cases, we randomly sample the initial conditions of the training trajectories, but the initial conditions of the test signals (of length ${N_{test}=100}$), which belong to both basins of attraction, form a ${50\times50}$ uniform grid. For all grid points, we plot the mean fraction correct calculated over ten random realizations of the RC's internal structure and of the initial conditions of the training signals. In \textbf{(a) and (b)}, we fix the number of training initial conditions, ${N_{train}=100}$, and in \textbf{(c) and (d)}, we fix the half-width of the test initial condition range, ${\Delta_0^{test}=1.5}$. Vertical lines mark the distance of the fixed points from the origin and diagonal lines mark ${\Delta_0^{test}=\Delta_0^{train}}$. Red crosses indicate the values of $N_{train}$, $\Delta_0^{train}$, and $\Delta_0^{test}$ that we use in \cref{fig:Magnetic_Basins,fig:Magnetic_Error_vs_Time,fig:Magnetic_Basins_TestLengths}.}
    \label{fig:Magnetic_Heatmaps}
\end{figure*}

\begin{figure*}[ht]
	\centering
	\includegraphics[width=.5\linewidth]{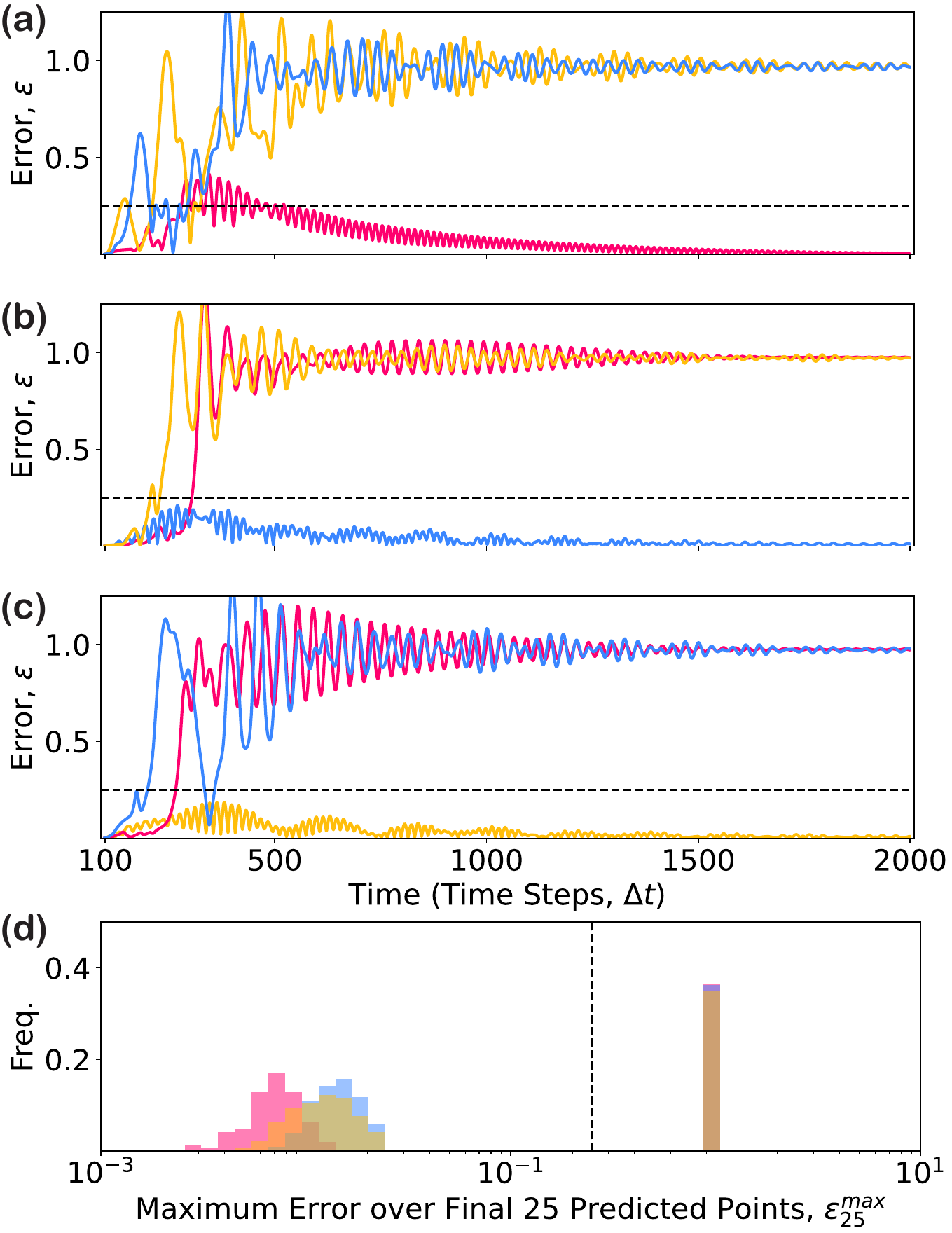}
	\caption{\justifying \textbf{RCs may learn some attractors more accurately than others, even as they generalize to unseen basins.} \textbf{(a)~to~(c)}~We measure the prediction error (Euclidean distance between the true and predicted trajectories) over time for sample predictions from short test signals (${N_{test}=100}$ data points) in the seen pink basin \textbf{(a)} and in the unseen blue \textbf{(b)} and yellow \textbf{(c)} basins. \textbf{(d)}~Distributions of the maximum errors over the final $25$ predicted points, calculated over $900$ predictions from test signals with randomly-distributed initial conditions ($310$ pink, $304$ blue, $286$ yellow). We color lines (a~to~c) according to their predicted basins and histograms (d) according to their true basins. In all panels, dashed lines indicate the threshold distance used to establish convergence, $\varepsilon_c=0.25$. We use the same RC and $N_{train}=100$ training trajectories from the pink basin of attraction as in \cref{fig:Magnetic_Basins}. The RC learns the dynamics near the seen pink attractor slightly more accurately than near the unseen blue and yellow attractors.
    }
    \label{fig:Magnetic_Error_vs_Time}
\end{figure*}

%% file: main.bbl
\begin{thebibliography}{62}%
\makeatletter
\providecommand \@ifxundefined [1]{%
 \@ifx{#1\undefined}
}%
\providecommand \@ifnum [1]{%
 \ifnum #1\expandafter \@firstoftwo
 \else \expandafter \@secondoftwo
 \fi
}%
\providecommand \@ifx [1]{%
 \ifx #1\expandafter \@firstoftwo
 \else \expandafter \@secondoftwo
 \fi
}%
\providecommand \natexlab [1]{#1}%
\providecommand \enquote  [1]{``#1''}%
\providecommand \bibnamefont  [1]{#1}%
\providecommand \bibfnamefont [1]{#1}%
\providecommand \citenamefont [1]{#1}%
\providecommand \href@noop [0]{\@secondoftwo}%
\providecommand \href [0]{\begingroup \@sanitize@url \@href}%
\providecommand \@href[1]{\@@startlink{#1}\@@href}%
\providecommand \@@href[1]{\endgroup#1\@@endlink}%
\providecommand \@sanitize@url [0]{\catcode `\\12\catcode `\$12\catcode `\&12\catcode `\#12\catcode `\^12\catcode `\_12\catcode `\%12\relax}%
\providecommand \@@startlink[1]{}%
\providecommand \@@endlink[0]{}%
\providecommand \url  [0]{\begingroup\@sanitize@url \@url }%
\providecommand \@url [1]{\endgroup\@href {#1}{\urlprefix }}%
\providecommand \urlprefix  [0]{URL }%
\providecommand \Eprint [0]{\href }%
\providecommand \doibase [0]{http://dx.doi.org/}%
\providecommand \selectlanguage [0]{\@gobble}%
\providecommand \bibinfo  [0]{\@secondoftwo}%
\providecommand \bibfield  [0]{\@secondoftwo}%
\providecommand \translation [1]{[#1]}%
\providecommand \BibitemOpen [0]{}%
\providecommand \bibitemStop [0]{}%
\providecommand \bibitemNoStop [0]{.\EOS\space}%
\providecommand \EOS [0]{\spacefactor3000\relax}%
\providecommand \BibitemShut  [1]{\csname bibitem#1\endcsname}%
\let\auto@bib@innerbib\@empty
%</preamble>
\bibitem [{\citenamefont {Brunton}\ and\ \citenamefont {Kutz}(2019)}]{Brunton2019_Book}%
  \BibitemOpen
  \bibfield  {author} {\bibinfo {author} {\bibfnamefont {S.~L.}\ \bibnamefont {Brunton}}\ and\ \bibinfo {author} {\bibfnamefont {J.~N.}\ \bibnamefont {Kutz}},\ }\href {\doibase 10.1017/9781108380690} {\emph {\bibinfo {title} {Data-Driven Science and Engineering: Machine Learning, Dynamical Systems, and Control}}}\ (\bibinfo  {publisher} {Cambridge University Press},\ \bibinfo {year} {2019})\BibitemShut {NoStop}%
\bibitem [{\citenamefont {Han}\ \emph {et~al.}(2021)\citenamefont {Han}, \citenamefont {Zhao}, \citenamefont {Leung}, \citenamefont {Ma},\ and\ \citenamefont {Wang}}]{Han2021_Deep_TS_Prediction_Review}%
  \BibitemOpen
  \bibfield  {author} {\bibinfo {author} {\bibfnamefont {Z.}~\bibnamefont {Han}}, \bibinfo {author} {\bibfnamefont {J.}~\bibnamefont {Zhao}}, \bibinfo {author} {\bibfnamefont {H.}~\bibnamefont {Leung}}, \bibinfo {author} {\bibfnamefont {K.~F.}\ \bibnamefont {Ma}}, \ and\ \bibinfo {author} {\bibfnamefont {W.}~\bibnamefont {Wang}},\ }\bibfield  {title} {\enquote {\bibinfo {title} {A review of deep learning models for time series prediction},}\ }\href {\doibase 10.1109/JSEN.2019.2923982} {\bibfield  {journal} {\bibinfo  {journal} {IEEE Sensors Journal}\ }\textbf {\bibinfo {volume} {21}},\ \bibinfo {pages} {7833--7848} (\bibinfo {year} {2021})}\BibitemShut {NoStop}%
\bibitem [{\citenamefont {Zhang}\ and\ \citenamefont {Cornelius}(2023)}]{Zhang2023_Catch22}%
  \BibitemOpen
  \bibfield  {author} {\bibinfo {author} {\bibfnamefont {Y.}~\bibnamefont {Zhang}}\ and\ \bibinfo {author} {\bibfnamefont {S.~P.}\ \bibnamefont {Cornelius}},\ }\bibfield  {title} {\enquote {\bibinfo {title} {Catch-22s of reservoir computing},}\ }\href {\doibase 10.1103/PhysRevResearch.5.033213} {\bibfield  {journal} {\bibinfo  {journal} {Phys. Rev. Res.}\ }\textbf {\bibinfo {volume} {5}},\ \bibinfo {pages} {033213} (\bibinfo {year} {2023})}\BibitemShut {NoStop}%
\bibitem [{\citenamefont {Göring}\ \emph {et~al.}(2024)\citenamefont {Göring}, \citenamefont {Hess}, \citenamefont {Brenner}, \citenamefont {Monfared},\ and\ \citenamefont {Durstewitz}}]{Goring2024_OutOfDomain}%
  \BibitemOpen
  \bibfield  {author} {\bibinfo {author} {\bibfnamefont {N.}~\bibnamefont {Göring}}, \bibinfo {author} {\bibfnamefont {F.}~\bibnamefont {Hess}}, \bibinfo {author} {\bibfnamefont {M.}~\bibnamefont {Brenner}}, \bibinfo {author} {\bibfnamefont {Z.}~\bibnamefont {Monfared}}, \ and\ \bibinfo {author} {\bibfnamefont {D.}~\bibnamefont {Durstewitz}},\ }\href@noop {} {\enquote {\bibinfo {title} {Out-of-domain generalization in dynamical systems reconstruction},}\ } (\bibinfo {year} {2024}),\ \Eprint {http://arxiv.org/abs/2402.18377} {arXiv:2402.18377 [cs.LG]} \BibitemShut {NoStop}%
\bibitem [{\citenamefont {Gauthier}, \citenamefont {Fischer},\ and\ \citenamefont {Röhm}(2022)}]{Gauthier2022_LearningUnseenCoexistingAttractors}%
  \BibitemOpen
  \bibfield  {author} {\bibinfo {author} {\bibfnamefont {D.~J.}\ \bibnamefont {Gauthier}}, \bibinfo {author} {\bibfnamefont {I.}~\bibnamefont {Fischer}}, \ and\ \bibinfo {author} {\bibfnamefont {A.}~\bibnamefont {Röhm}},\ }\bibfield  {title} {\enquote {\bibinfo {title} {Learning unseen coexisting attractors},}\ }\href {\doibase 10.1063/5.0116784} {\bibfield  {journal} {\bibinfo  {journal} {Chaos: An Interdisciplinary Journal of Nonlinear Science}\ }\textbf {\bibinfo {volume} {32}},\ \bibinfo {pages} {113107} (\bibinfo {year} {2022})}\BibitemShut {NoStop}%
\bibitem [{\citenamefont {Yu}\ and\ \citenamefont {Wang}(2024)}]{Yu2024_PhysicsGuidedDL}%
  \BibitemOpen
  \bibfield  {author} {\bibinfo {author} {\bibfnamefont {R.}~\bibnamefont {Yu}}\ and\ \bibinfo {author} {\bibfnamefont {R.}~\bibnamefont {Wang}},\ }\bibfield  {title} {\enquote {\bibinfo {title} {Learning dynamical systems from data: An introduction to physics-guided deep learning},}\ }\href {\doibase 10.1073/pnas.2311808121} {\bibfield  {journal} {\bibinfo  {journal} {Proceedings of the National Academy of Sciences}\ }\textbf {\bibinfo {volume} {121}},\ \bibinfo {pages} {e2311808121} (\bibinfo {year} {2024})}\BibitemShut {NoStop}%
\bibitem [{\citenamefont {Brunton}, \citenamefont {Proctor},\ and\ \citenamefont {Kutz}(2016)}]{Brunton2016_SINDy}%
  \BibitemOpen
  \bibfield  {author} {\bibinfo {author} {\bibfnamefont {S.~L.}\ \bibnamefont {Brunton}}, \bibinfo {author} {\bibfnamefont {J.~L.}\ \bibnamefont {Proctor}}, \ and\ \bibinfo {author} {\bibfnamefont {J.~N.}\ \bibnamefont {Kutz}},\ }\bibfield  {title} {\enquote {\bibinfo {title} {Discovering governing equations from data by sparse identification of nonlinear dynamical systems},}\ }\href {\doibase 10.1073/pnas.1517384113} {\bibfield  {journal} {\bibinfo  {journal} {Proceedings of the National Academy of Sciences}\ }\textbf {\bibinfo {volume} {113}},\ \bibinfo {pages} {3932--3937} (\bibinfo {year} {2016})}\BibitemShut {NoStop}%
\bibitem [{\citenamefont {Rudy}\ \emph {et~al.}(2017)\citenamefont {Rudy}, \citenamefont {Brunton}, \citenamefont {Proctor},\ and\ \citenamefont {Kutz}}]{Rudy2017_PDESindy}%
  \BibitemOpen
  \bibfield  {author} {\bibinfo {author} {\bibfnamefont {S.~H.}\ \bibnamefont {Rudy}}, \bibinfo {author} {\bibfnamefont {S.~L.}\ \bibnamefont {Brunton}}, \bibinfo {author} {\bibfnamefont {J.~L.}\ \bibnamefont {Proctor}}, \ and\ \bibinfo {author} {\bibfnamefont {J.~N.}\ \bibnamefont {Kutz}},\ }\bibfield  {title} {\enquote {\bibinfo {title} {Data-driven discovery of partial differential equations},}\ }\href {\doibase 10.1126/sciadv.1602614} {\bibfield  {journal} {\bibinfo  {journal} {Science Advances}\ }\textbf {\bibinfo {volume} {3}},\ \bibinfo {pages} {e1602614} (\bibinfo {year} {2017})}\BibitemShut {NoStop}%
\bibitem [{\citenamefont {Gauthier}\ \emph {et~al.}(2021)\citenamefont {Gauthier}, \citenamefont {Bollt}, \citenamefont {Griffith},\ and\ \citenamefont {Barbosa}}]{Gauthier2021_NGRC}%
  \BibitemOpen
  \bibfield  {author} {\bibinfo {author} {\bibfnamefont {D.~J.}\ \bibnamefont {Gauthier}}, \bibinfo {author} {\bibfnamefont {E.}~\bibnamefont {Bollt}}, \bibinfo {author} {\bibfnamefont {A.}~\bibnamefont {Griffith}}, \ and\ \bibinfo {author} {\bibfnamefont {W.~A.~S.}\ \bibnamefont {Barbosa}},\ }\bibfield  {title} {\enquote {\bibinfo {title} {Next generation reservoir computing},}\ }\href {\doibase 10.1038/s41467-021-25801-2} {\bibfield  {journal} {\bibinfo  {journal} {Nature Communications}\ }\textbf {\bibinfo {volume} {12}} (\bibinfo {year} {2021}),\ 10.1038/s41467-021-25801-2}\BibitemShut {NoStop}%
\bibitem [{\citenamefont {Pathak}\ \emph {et~al.}(2018{\natexlab{a}})\citenamefont {Pathak}, \citenamefont {Wikner}, \citenamefont {Fussell}, \citenamefont {Chandra}, \citenamefont {Hunt}, \citenamefont {Girvan},\ and\ \citenamefont {Ott}}]{Pathak_Hybrid}%
  \BibitemOpen
  \bibfield  {author} {\bibinfo {author} {\bibfnamefont {J.}~\bibnamefont {Pathak}}, \bibinfo {author} {\bibfnamefont {A.}~\bibnamefont {Wikner}}, \bibinfo {author} {\bibfnamefont {R.}~\bibnamefont {Fussell}}, \bibinfo {author} {\bibfnamefont {S.}~\bibnamefont {Chandra}}, \bibinfo {author} {\bibfnamefont {B.~R.}\ \bibnamefont {Hunt}}, \bibinfo {author} {\bibfnamefont {M.}~\bibnamefont {Girvan}}, \ and\ \bibinfo {author} {\bibfnamefont {E.}~\bibnamefont {Ott}},\ }\bibfield  {title} {\enquote {\bibinfo {title} {{Hybrid forecasting of chaotic processes: Using machine learning in conjunction with a knowledge-based model}},}\ }\href {\doibase 10.1063/1.5028373} {\bibfield  {journal} {\bibinfo  {journal} {Chaos: An Interdisciplinary Journal of Nonlinear Science}\ }\textbf {\bibinfo {volume} {28}},\ \bibinfo {pages} {041101} (\bibinfo {year} {2018}{\natexlab{a}})}\BibitemShut {NoStop}%
\bibitem [{\citenamefont {Arcomano}\ \emph {et~al.}(2022)\citenamefont {Arcomano}, \citenamefont {Szunyogh}, \citenamefont {Wikner}, \citenamefont {Pathak}, \citenamefont {Hunt},\ and\ \citenamefont {Ott}}]{Arcomano2022_AtmHybrid}%
  \BibitemOpen
  \bibfield  {author} {\bibinfo {author} {\bibfnamefont {T.}~\bibnamefont {Arcomano}}, \bibinfo {author} {\bibfnamefont {I.}~\bibnamefont {Szunyogh}}, \bibinfo {author} {\bibfnamefont {A.}~\bibnamefont {Wikner}}, \bibinfo {author} {\bibfnamefont {J.}~\bibnamefont {Pathak}}, \bibinfo {author} {\bibfnamefont {B.~R.}\ \bibnamefont {Hunt}}, \ and\ \bibinfo {author} {\bibfnamefont {E.}~\bibnamefont {Ott}},\ }\bibfield  {title} {\enquote {\bibinfo {title} {A hybrid approach to atmospheric modeling that combines machine learning with a physics-based numerical model},}\ }\href {\doibase https://doi.org/10.1029/2021MS002712} {\bibfield  {journal} {\bibinfo  {journal} {Journal of Advances in Modeling Earth Systems}\ }\textbf {\bibinfo {volume} {14}},\ \bibinfo {pages} {e2021MS002712} (\bibinfo {year} {2022})}\BibitemShut {NoStop}%
\bibitem [{\citenamefont {Chepuri}\ \emph {et~al.}(2024)\citenamefont {Chepuri}, \citenamefont {Amzalag}, \citenamefont {Antonsen},\ and\ \citenamefont {Girvan}}]{Chepuri2024_RC_NGRC_Hybrid}%
  \BibitemOpen
  \bibfield  {author} {\bibinfo {author} {\bibfnamefont {R.}~\bibnamefont {Chepuri}}, \bibinfo {author} {\bibfnamefont {D.}~\bibnamefont {Amzalag}}, \bibinfo {author} {\bibfnamefont {T.~M.}\ \bibnamefont {Antonsen}}, \ and\ \bibinfo {author} {\bibfnamefont {M.}~\bibnamefont {Girvan}},\ }\bibfield  {title} {\enquote {\bibinfo {title} {Hybridizing traditional and next-generation reservoir computing to accurately and efficiently forecast dynamical systems},}\ }\href {\doibase 10.1063/5.0206232} {\bibfield  {journal} {\bibinfo  {journal} {Chaos: An Interdisciplinary Journal of Nonlinear Science}\ }\textbf {\bibinfo {volume} {34}},\ \bibinfo {pages} {063114} (\bibinfo {year} {2024})}\BibitemShut {NoStop}%
\bibitem [{\citenamefont {Vardi}(2023)}]{Vardi2023_ImplicitBias}%
  \BibitemOpen
  \bibfield  {author} {\bibinfo {author} {\bibfnamefont {G.}~\bibnamefont {Vardi}},\ }\bibfield  {title} {\enquote {\bibinfo {title} {On the implicit bias in deep-learning algorithms},}\ }\href {\doibase 10.1145/3571070} {\bibfield  {journal} {\bibinfo  {journal} {Commun. ACM}\ }\textbf {\bibinfo {volume} {66}},\ \bibinfo {pages} {86–93} (\bibinfo {year} {2023})}\BibitemShut {NoStop}%
\bibitem [{\citenamefont {Ribeiro}\ \emph {et~al.}(2021)\citenamefont {Ribeiro}, \citenamefont {Hendriks}, \citenamefont {Wills},\ and\ \citenamefont {Schön}}]{Ribeiro2021_DoubleDescentModelingDynamics}%
  \BibitemOpen
  \bibfield  {author} {\bibinfo {author} {\bibfnamefont {A.~H.}\ \bibnamefont {Ribeiro}}, \bibinfo {author} {\bibfnamefont {J.~N.}\ \bibnamefont {Hendriks}}, \bibinfo {author} {\bibfnamefont {A.~G.}\ \bibnamefont {Wills}}, \ and\ \bibinfo {author} {\bibfnamefont {T.~B.}\ \bibnamefont {Schön}},\ }\bibfield  {title} {\enquote {\bibinfo {title} {Beyond occam’s razor in system identification: Double-descent when modeling dynamics},}\ }\href {\doibase https://doi.org/10.1016/j.ifacol.2021.08.341} {\bibfield  {journal} {\bibinfo  {journal} {IFAC-PapersOnLine}\ }\textbf {\bibinfo {volume} {54}},\ \bibinfo {pages} {97--102} (\bibinfo {year} {2021})},\ \bibinfo {note} {19th IFAC Symposium on System Identification SYSID 2021}\BibitemShut {NoStop}%
\bibitem [{\citenamefont {Röhm}, \citenamefont {Gauthier},\ and\ \citenamefont {Fischer}(2021)}]{Rohm2021_UnseenAttractors_RC}%
  \BibitemOpen
  \bibfield  {author} {\bibinfo {author} {\bibfnamefont {A.}~\bibnamefont {Röhm}}, \bibinfo {author} {\bibfnamefont {D.~J.}\ \bibnamefont {Gauthier}}, \ and\ \bibinfo {author} {\bibfnamefont {I.}~\bibnamefont {Fischer}},\ }\bibfield  {title} {\enquote {\bibinfo {title} {{Model-free inference of unseen attractors: Reconstructing phase space features from a single noisy trajectory using reservoir computing}},}\ }\href {\doibase 10.1063/5.0065813} {\bibfield  {journal} {\bibinfo  {journal} {Chaos: An Interdisciplinary Journal of Nonlinear Science}\ }\textbf {\bibinfo {volume} {31}},\ \bibinfo {pages} {103127} (\bibinfo {year} {2021})}\BibitemShut {NoStop}%
\bibitem [{\citenamefont {Du}\ \emph {et~al.}(2024)\citenamefont {Du}, \citenamefont {Li}, \citenamefont {Fan}, \citenamefont {Zhan}, \citenamefont {Xiao},\ and\ \citenamefont {Wang}}]{Du2024_PowerSystemBasinsML}%
  \BibitemOpen
  \bibfield  {author} {\bibinfo {author} {\bibfnamefont {Y.}~\bibnamefont {Du}}, \bibinfo {author} {\bibfnamefont {Q.}~\bibnamefont {Li}}, \bibinfo {author} {\bibfnamefont {H.}~\bibnamefont {Fan}}, \bibinfo {author} {\bibfnamefont {M.}~\bibnamefont {Zhan}}, \bibinfo {author} {\bibfnamefont {J.}~\bibnamefont {Xiao}}, \ and\ \bibinfo {author} {\bibfnamefont {X.}~\bibnamefont {Wang}},\ }\bibfield  {title} {\enquote {\bibinfo {title} {Inferring attracting basins of power system with machine learning},}\ }\href {\doibase 10.1103/PhysRevResearch.6.013181} {\bibfield  {journal} {\bibinfo  {journal} {Phys. Rev. Res.}\ }\textbf {\bibinfo {volume} {6}},\ \bibinfo {pages} {013181} (\bibinfo {year} {2024})}\BibitemShut {NoStop}%
\bibitem [{\citenamefont {Jaeger}\ and\ \citenamefont {Haas}(2004)}]{JaegarHaas}%
  \BibitemOpen
  \bibfield  {author} {\bibinfo {author} {\bibfnamefont {H.}~\bibnamefont {Jaeger}}\ and\ \bibinfo {author} {\bibfnamefont {H.}~\bibnamefont {Haas}},\ }\bibfield  {title} {\enquote {\bibinfo {title} {Harnessing nonlinearity: Predicting chaotic systems and saving energy in wireless communication},}\ }\href {\doibase 10.1126/science.1091277} {\bibfield  {journal} {\bibinfo  {journal} {Science}\ }\textbf {\bibinfo {volume} {304}},\ \bibinfo {pages} {78--80} (\bibinfo {year} {2004})}\BibitemShut {NoStop}%
\bibitem [{\citenamefont {Schrauwen}, \citenamefont {Verstraeten},\ and\ \citenamefont {Campenhout}(2007)}]{ESANN_Res_Overview}%
  \BibitemOpen
  \bibfield  {author} {\bibinfo {author} {\bibfnamefont {B.}~\bibnamefont {Schrauwen}}, \bibinfo {author} {\bibfnamefont {D.}~\bibnamefont {Verstraeten}}, \ and\ \bibinfo {author} {\bibfnamefont {J.}~\bibnamefont {Campenhout}},\ }\bibfield  {title} {\enquote {\bibinfo {title} {An overview of reservoir computing: Theory, applications and implementations},}\ \ }(\bibinfo {year} {2007})\ pp.\ \bibinfo {pages} {471--482}\BibitemShut {NoStop}%
\bibitem [{\citenamefont {Sun}\ \emph {et~al.}(2024)\citenamefont {Sun}, \citenamefont {Song}, \citenamefont {Cai}, \citenamefont {Zhang}, \citenamefont {Hong},\ and\ \citenamefont {Li}}]{Sun2024_ESN_Review}%
  \BibitemOpen
  \bibfield  {author} {\bibinfo {author} {\bibfnamefont {C.}~\bibnamefont {Sun}}, \bibinfo {author} {\bibfnamefont {M.}~\bibnamefont {Song}}, \bibinfo {author} {\bibfnamefont {D.}~\bibnamefont {Cai}}, \bibinfo {author} {\bibfnamefont {B.}~\bibnamefont {Zhang}}, \bibinfo {author} {\bibfnamefont {S.}~\bibnamefont {Hong}}, \ and\ \bibinfo {author} {\bibfnamefont {H.}~\bibnamefont {Li}},\ }\bibfield  {title} {\enquote {\bibinfo {title} {A systematic review of echo state networks from design to application},}\ }\href {\doibase 10.1109/TAI.2022.3225780} {\bibfield  {journal} {\bibinfo  {journal} {IEEE Transactions on Artificial Intelligence}\ }\textbf {\bibinfo {volume} {5}},\ \bibinfo {pages} {23--37} (\bibinfo {year} {2024})}\BibitemShut {NoStop}%
\bibitem [{\citenamefont {Lukoševičius}\ and\ \citenamefont {Jaeger}(2009)}]{Lukosevicius2009_RC_Review}%
  \BibitemOpen
  \bibfield  {author} {\bibinfo {author} {\bibfnamefont {M.}~\bibnamefont {Lukoševičius}}\ and\ \bibinfo {author} {\bibfnamefont {H.}~\bibnamefont {Jaeger}},\ }\bibfield  {title} {\enquote {\bibinfo {title} {Reservoir computing approaches to recurrent neural network training},}\ }\href {\doibase https://doi.org/10.1016/j.cosrev.2009.03.005} {\bibfield  {journal} {\bibinfo  {journal} {Computer Science Review}\ }\textbf {\bibinfo {volume} {3}},\ \bibinfo {pages} {127--149} (\bibinfo {year} {2009})}\BibitemShut {NoStop}%
\bibitem [{\citenamefont {Lu}\ \emph {et~al.}(2017)\citenamefont {Lu}, \citenamefont {Pathak}, \citenamefont {Hunt}, \citenamefont {Girvan}, \citenamefont {Brockett},\ and\ \citenamefont {Ott}}]{Lu_and_Pathak}%
  \BibitemOpen
  \bibfield  {author} {\bibinfo {author} {\bibfnamefont {Z.}~\bibnamefont {Lu}}, \bibinfo {author} {\bibfnamefont {J.}~\bibnamefont {Pathak}}, \bibinfo {author} {\bibfnamefont {B.}~\bibnamefont {Hunt}}, \bibinfo {author} {\bibfnamefont {M.}~\bibnamefont {Girvan}}, \bibinfo {author} {\bibfnamefont {R.}~\bibnamefont {Brockett}}, \ and\ \bibinfo {author} {\bibfnamefont {E.}~\bibnamefont {Ott}},\ }\bibfield  {title} {\enquote {\bibinfo {title} {Reservoir observers: Model-free inference of unmeasured variables in chaotic systems},}\ }\href {\doibase 10.1063/1.4979665} {\bibfield  {journal} {\bibinfo  {journal} {Chaos: An Interdisciplinary Journal of Nonlinear Science}\ }\textbf {\bibinfo {volume} {27}},\ \bibinfo {pages} {041102} (\bibinfo {year} {2017})}\BibitemShut {NoStop}%
\bibitem [{\citenamefont {Srinivasan}\ \emph {et~al.}(2022)\citenamefont {Srinivasan}, \citenamefont {Coble}, \citenamefont {Hamlin}, \citenamefont {Antonsen}, \citenamefont {Ott},\ and\ \citenamefont {Girvan}}]{Srinivasan2022_ParallelRC_for_Networks}%
  \BibitemOpen
  \bibfield  {author} {\bibinfo {author} {\bibfnamefont {K.}~\bibnamefont {Srinivasan}}, \bibinfo {author} {\bibfnamefont {N.}~\bibnamefont {Coble}}, \bibinfo {author} {\bibfnamefont {J.}~\bibnamefont {Hamlin}}, \bibinfo {author} {\bibfnamefont {T.}~\bibnamefont {Antonsen}}, \bibinfo {author} {\bibfnamefont {E.}~\bibnamefont {Ott}}, \ and\ \bibinfo {author} {\bibfnamefont {M.}~\bibnamefont {Girvan}},\ }\bibfield  {title} {\enquote {\bibinfo {title} {Parallel machine learning for forecasting the dynamics of complex networks},}\ }\href {\doibase 10.1103/PhysRevLett.128.164101} {\bibfield  {journal} {\bibinfo  {journal} {Phys. Rev. Lett.}\ }\textbf {\bibinfo {volume} {128}},\ \bibinfo {pages} {164101} (\bibinfo {year} {2022})}\BibitemShut {NoStop}%
\bibitem [{\citenamefont {Pathak}\ \emph {et~al.}(2018{\natexlab{b}})\citenamefont {Pathak}, \citenamefont {Hunt}, \citenamefont {Girvan}, \citenamefont {Lu},\ and\ \citenamefont {Ott}}]{Pathak2018_SpatioTemporalChaosRC}%
  \BibitemOpen
  \bibfield  {author} {\bibinfo {author} {\bibfnamefont {J.}~\bibnamefont {Pathak}}, \bibinfo {author} {\bibfnamefont {B.}~\bibnamefont {Hunt}}, \bibinfo {author} {\bibfnamefont {M.}~\bibnamefont {Girvan}}, \bibinfo {author} {\bibfnamefont {Z.}~\bibnamefont {Lu}}, \ and\ \bibinfo {author} {\bibfnamefont {E.}~\bibnamefont {Ott}},\ }\bibfield  {title} {\enquote {\bibinfo {title} {Model-free prediction of large spatiotemporally chaotic systems from data: A reservoir computing approach},}\ }\href {\doibase 10.1103/PhysRevLett.120.024102} {\bibfield  {journal} {\bibinfo  {journal} {Phys. Rev. Lett.}\ }\textbf {\bibinfo {volume} {120}},\ \bibinfo {pages} {024102} (\bibinfo {year} {2018}{\natexlab{b}})}\BibitemShut {NoStop}%
\bibitem [{\citenamefont {Krishnagopal}\ \emph {et~al.}(2020)\citenamefont {Krishnagopal}, \citenamefont {Girvan}, \citenamefont {Ott},\ and\ \citenamefont {Hunt}}]{Krishnagopal2020_Sep_of_Chaotic_Signals}%
  \BibitemOpen
  \bibfield  {author} {\bibinfo {author} {\bibfnamefont {S.}~\bibnamefont {Krishnagopal}}, \bibinfo {author} {\bibfnamefont {M.}~\bibnamefont {Girvan}}, \bibinfo {author} {\bibfnamefont {E.}~\bibnamefont {Ott}}, \ and\ \bibinfo {author} {\bibfnamefont {B.~R.}\ \bibnamefont {Hunt}},\ }\bibfield  {title} {\enquote {\bibinfo {title} {{Separation of chaotic signals by reservoir computing}},}\ }\href {\doibase 10.1063/1.5132766} {\bibfield  {journal} {\bibinfo  {journal} {Chaos: An Interdisciplinary Journal of Nonlinear Science}\ }\textbf {\bibinfo {volume} {30}},\ \bibinfo {pages} {023123} (\bibinfo {year} {2020})}\BibitemShut {NoStop}%
\bibitem [{\citenamefont {Banerjee}\ \emph {et~al.}(2021)\citenamefont {Banerjee}, \citenamefont {Hart}, \citenamefont {Roy},\ and\ \citenamefont {Ott}}]{Banerjee2021_RCLinkInference}%
  \BibitemOpen
  \bibfield  {author} {\bibinfo {author} {\bibfnamefont {A.}~\bibnamefont {Banerjee}}, \bibinfo {author} {\bibfnamefont {J.~D.}\ \bibnamefont {Hart}}, \bibinfo {author} {\bibfnamefont {R.}~\bibnamefont {Roy}}, \ and\ \bibinfo {author} {\bibfnamefont {E.}~\bibnamefont {Ott}},\ }\bibfield  {title} {\enquote {\bibinfo {title} {Machine learning link inference of noisy delay-coupled networks with optoelectronic experimental tests},}\ }\href {\doibase 10.1103/PhysRevX.11.031014} {\bibfield  {journal} {\bibinfo  {journal} {Phys. Rev. X}\ }\textbf {\bibinfo {volume} {11}},\ \bibinfo {pages} {031014} (\bibinfo {year} {2021})}\BibitemShut {NoStop}%
\bibitem [{\citenamefont {Tanaka}\ \emph {et~al.}(2019)\citenamefont {Tanaka} \emph {et~al.}}]{Tanaka_RC_Review_2019}%
  \BibitemOpen
  \bibfield  {author} {\bibinfo {author} {\bibfnamefont {G.}~\bibnamefont {Tanaka}} \emph {et~al.},\ }\bibfield  {title} {\enquote {\bibinfo {title} {Recent advances in physical reservoir computing: A review},}\ }\href {\doibase https://doi.org/10.1016/j.neunet.2019.03.005} {\bibfield  {journal} {\bibinfo  {journal} {Neural Networks}\ }\textbf {\bibinfo {volume} {115}},\ \bibinfo {pages} {100--123} (\bibinfo {year} {2019})}\BibitemShut {NoStop}%
\bibitem [{\citenamefont {Bollt}(2021)}]{Bollt_RC_VAR}%
  \BibitemOpen
  \bibfield  {author} {\bibinfo {author} {\bibfnamefont {E.}~\bibnamefont {Bollt}},\ }\bibfield  {title} {\enquote {\bibinfo {title} {On explaining the surprising success of reservoir computing forecaster of chaos? the universal machine learning dynamical system with contrast to var and dmd},}\ }\href {\doibase 10.1063/5.0024890} {\bibfield  {journal} {\bibinfo  {journal} {Chaos: An Interdisciplinary Journal of Nonlinear Science}\ }\textbf {\bibinfo {volume} {31}},\ \bibinfo {pages} {013108} (\bibinfo {year} {2021})}\BibitemShut {NoStop}%
\bibitem [{\citenamefont {Wagemakers}(2025)}]{Wagemakers2025_BasinZoo}%
  \BibitemOpen
  \bibfield  {author} {\bibinfo {author} {\bibfnamefont {A.}~\bibnamefont {Wagemakers}},\ }\href {https://arxiv.org/abs/2504.01580} {\enquote {\bibinfo {title} {The basins zoo},}\ } (\bibinfo {year} {2025}),\ \Eprint {http://arxiv.org/abs/2504.01580} {arXiv:2504.01580 [nlin.CD]} \BibitemShut {NoStop}%
\bibitem [{\citenamefont {Izhikevich}(2006)}]{Izhikevich2006_DynSysNeuroscience}%
  \BibitemOpen
  \bibfield  {author} {\bibinfo {author} {\bibfnamefont {E.~M.}\ \bibnamefont {Izhikevich}},\ }\href {\doibase 10.7551/mitpress/2526.001.0001} {\emph {\bibinfo {title} {Dynamical Systems in Neuroscience: The Geometry of Excitability and Bursting}}}\ (\bibinfo  {publisher} {The MIT Press},\ \bibinfo {year} {2006})\BibitemShut {NoStop}%
\bibitem [{\citenamefont {Rand}\ \emph {et~al.}(2021)\citenamefont {Rand}, \citenamefont {Raju}, \citenamefont {Sáez}, \citenamefont {Corson},\ and\ \citenamefont {Siggia}}]{Rand2021_GeneRegulatoryDynamics}%
  \BibitemOpen
  \bibfield  {author} {\bibinfo {author} {\bibfnamefont {D.~A.}\ \bibnamefont {Rand}}, \bibinfo {author} {\bibfnamefont {A.}~\bibnamefont {Raju}}, \bibinfo {author} {\bibfnamefont {M.}~\bibnamefont {Sáez}}, \bibinfo {author} {\bibfnamefont {F.}~\bibnamefont {Corson}}, \ and\ \bibinfo {author} {\bibfnamefont {E.~D.}\ \bibnamefont {Siggia}},\ }\bibfield  {title} {\enquote {\bibinfo {title} {Geometry of gene regulatory dynamics},}\ }\href {\doibase 10.1073/pnas.2109729118} {\bibfield  {journal} {\bibinfo  {journal} {Proceedings of the National Academy of Sciences}\ }\textbf {\bibinfo {volume} {118}},\ \bibinfo {pages} {e2109729118} (\bibinfo {year} {2021})}\BibitemShut {NoStop}%
\bibitem [{\citenamefont {Corson}\ \emph {et~al.}(2017)\citenamefont {Corson}, \citenamefont {Couturier}, \citenamefont {Rouault}, \citenamefont {Mazouni},\ and\ \citenamefont {Schweisguth}}]{Corson2017_Drosophila}%
  \BibitemOpen
  \bibfield  {author} {\bibinfo {author} {\bibfnamefont {F.}~\bibnamefont {Corson}}, \bibinfo {author} {\bibfnamefont {L.}~\bibnamefont {Couturier}}, \bibinfo {author} {\bibfnamefont {H.}~\bibnamefont {Rouault}}, \bibinfo {author} {\bibfnamefont {K.}~\bibnamefont {Mazouni}}, \ and\ \bibinfo {author} {\bibfnamefont {F.}~\bibnamefont {Schweisguth}},\ }\bibfield  {title} {\enquote {\bibinfo {title} {Self-organized notch dynamics generate stereotyped sensory organ patterns in drosophila},}\ }\href {\doibase 10.1126/science.aai7407} {\bibfield  {journal} {\bibinfo  {journal} {Science}\ }\textbf {\bibinfo {volume} {356}},\ \bibinfo {pages} {eaai7407} (\bibinfo {year} {2017})}\BibitemShut {NoStop}%
\bibitem [{\citenamefont {Menck}\ \emph {et~al.}(2014)\citenamefont {Menck}, \citenamefont {Heitzig}, \citenamefont {Kurths},\ and\ \citenamefont {Joachim~Schellnhuber}}]{Menck2014_PowerGridStability}%
  \BibitemOpen
  \bibfield  {author} {\bibinfo {author} {\bibfnamefont {P.~J.}\ \bibnamefont {Menck}}, \bibinfo {author} {\bibfnamefont {J.}~\bibnamefont {Heitzig}}, \bibinfo {author} {\bibfnamefont {J.}~\bibnamefont {Kurths}}, \ and\ \bibinfo {author} {\bibfnamefont {H.}~\bibnamefont {Joachim~Schellnhuber}},\ }\bibfield  {title} {\enquote {\bibinfo {title} {How dead ends undermine power grid stability},}\ }\href {\doibase 10.1038/ncomms4969} {\bibfield  {journal} {\bibinfo  {journal} {Nature Communications}\ }\textbf {\bibinfo {volume} {5}},\ \bibinfo {pages} {3969} (\bibinfo {year} {2014})}\BibitemShut {NoStop}%
\bibitem [{\citenamefont {Cavalli}\ and\ \citenamefont {Naimzada}(2016)}]{Cavalli2016_MultistabilityMarketGames}%
  \BibitemOpen
  \bibfield  {author} {\bibinfo {author} {\bibfnamefont {F.}~\bibnamefont {Cavalli}}\ and\ \bibinfo {author} {\bibfnamefont {A.}~\bibnamefont {Naimzada}},\ }\bibfield  {title} {\enquote {\bibinfo {title} {Complex dynamics and multistability with increasing rationality in market games},}\ }\href {\doibase https://doi.org/10.1016/j.chaos.2016.10.014} {\bibfield  {journal} {\bibinfo  {journal} {Chaos, Solitons \& Fractals}\ }\textbf {\bibinfo {volume} {93}},\ \bibinfo {pages} {151--161} (\bibinfo {year} {2016})}\BibitemShut {NoStop}%
\bibitem [{\citenamefont {Norton}\ \emph {et~al.}(2025)\citenamefont {Norton}, \citenamefont {Ott}, \citenamefont {Pomerance}, \citenamefont {Hunt},\ and\ \citenamefont {Girvan}}]{Norton2025_METAFORS}%
  \BibitemOpen
  \bibfield  {author} {\bibinfo {author} {\bibfnamefont {D.~A.}\ \bibnamefont {Norton}}, \bibinfo {author} {\bibfnamefont {E.}~\bibnamefont {Ott}}, \bibinfo {author} {\bibfnamefont {A.}~\bibnamefont {Pomerance}}, \bibinfo {author} {\bibfnamefont {B.}~\bibnamefont {Hunt}}, \ and\ \bibinfo {author} {\bibfnamefont {M.}~\bibnamefont {Girvan}},\ }\href {https://arxiv.org/abs/2501.16325} {\enquote {\bibinfo {title} {Tailored forecasting from short time series via meta-learning},}\ } (\bibinfo {year} {2025}),\ \Eprint {http://arxiv.org/abs/2501.16325} {arXiv:2501.16325 [cs.LG]} \BibitemShut {NoStop}%
\bibitem [{\citenamefont {Kong}\ \emph {et~al.}(2021)\citenamefont {Kong}, \citenamefont {Fan}, \citenamefont {Grebogi},\ and\ \citenamefont {Lai}}]{Kong2021_MLSystemCollapse}%
  \BibitemOpen
  \bibfield  {author} {\bibinfo {author} {\bibfnamefont {L.-W.}\ \bibnamefont {Kong}}, \bibinfo {author} {\bibfnamefont {H.-W.}\ \bibnamefont {Fan}}, \bibinfo {author} {\bibfnamefont {C.}~\bibnamefont {Grebogi}}, \ and\ \bibinfo {author} {\bibfnamefont {Y.-C.}\ \bibnamefont {Lai}},\ }\bibfield  {title} {\enquote {\bibinfo {title} {Machine learning prediction of critical transition and system collapse},}\ }\href {\doibase 10.1103/PhysRevResearch.3.013090} {\bibfield  {journal} {\bibinfo  {journal} {Phys. Rev. Res.}\ }\textbf {\bibinfo {volume} {3}},\ \bibinfo {pages} {013090} (\bibinfo {year} {2021})}\BibitemShut {NoStop}%
\bibitem [{\citenamefont {Panahi}\ and\ \citenamefont {Lai}(2024)}]{Panahi2024_AdaptableRC}%
  \BibitemOpen
  \bibfield  {author} {\bibinfo {author} {\bibfnamefont {S.}~\bibnamefont {Panahi}}\ and\ \bibinfo {author} {\bibfnamefont {Y.-C.}\ \bibnamefont {Lai}},\ }\bibfield  {title} {\enquote {\bibinfo {title} {{Adaptable reservoir computing: A paradigm for model-free data-driven prediction of critical transitions in nonlinear dynamical systems}},}\ }\href {\doibase 10.1063/5.0200898} {\bibfield  {journal} {\bibinfo  {journal} {Chaos: An Interdisciplinary Journal of Nonlinear Science}\ }\textbf {\bibinfo {volume} {34}},\ \bibinfo {pages} {051501} (\bibinfo {year} {2024})}\BibitemShut {NoStop}%
\bibitem [{\citenamefont {Kong}, \citenamefont {Brewer},\ and\ \citenamefont {Lai}(2024)}]{Kong2024_IndexBasedRC}%
  \BibitemOpen
  \bibfield  {author} {\bibinfo {author} {\bibfnamefont {L.-W.}\ \bibnamefont {Kong}}, \bibinfo {author} {\bibfnamefont {G.~A.}\ \bibnamefont {Brewer}}, \ and\ \bibinfo {author} {\bibfnamefont {Y.-C.}\ \bibnamefont {Lai}},\ }\bibfield  {title} {\enquote {\bibinfo {title} {Reservoir-computing based associative memory and itinerancy for complex dynamical attractors},}\ }\href {\doibase 10.1038/s41467-024-49190-4} {\bibfield  {journal} {\bibinfo  {journal} {Nature Communications}\ }\textbf {\bibinfo {volume} {15}},\ \bibinfo {pages} {4840} (\bibinfo {year} {2024})}\BibitemShut {NoStop}%
\bibitem [{\citenamefont {Kim}\ \emph {et~al.}(2020)\citenamefont {Kim}, \citenamefont {Lu}, \citenamefont {Nozari}, \citenamefont {Pappas},\ and\ \citenamefont {Bassett}}]{Kim2020_RNNChaoticMemories}%
  \BibitemOpen
  \bibfield  {author} {\bibinfo {author} {\bibfnamefont {J.~Z.}\ \bibnamefont {Kim}}, \bibinfo {author} {\bibfnamefont {Z.}~\bibnamefont {Lu}}, \bibinfo {author} {\bibfnamefont {E.}~\bibnamefont {Nozari}}, \bibinfo {author} {\bibfnamefont {G.~J.}\ \bibnamefont {Pappas}}, \ and\ \bibinfo {author} {\bibfnamefont {D.~S.}\ \bibnamefont {Bassett}},\ }\href {https://arxiv.org/abs/2005.01186} {\enquote {\bibinfo {title} {Teaching recurrent neural networks to modify chaotic memories by example},}\ } (\bibinfo {year} {2020}),\ \Eprint {http://arxiv.org/abs/2005.01186} {arXiv:2005.01186 [cond-mat.dis-nn]} \BibitemShut {NoStop}%
\bibitem [{\citenamefont {Lu}\ and\ \citenamefont {Bassett}(2020)}]{Lu2020_IGS}%
  \BibitemOpen
  \bibfield  {author} {\bibinfo {author} {\bibfnamefont {Z.}~\bibnamefont {Lu}}\ and\ \bibinfo {author} {\bibfnamefont {D.~S.}\ \bibnamefont {Bassett}},\ }\bibfield  {title} {\enquote {\bibinfo {title} {{Invertible generalized synchronization: A putative mechanism for implicit learning in neural systems}},}\ }\href {\doibase 10.1063/5.0004344} {\bibfield  {journal} {\bibinfo  {journal} {Chaos: An Interdisciplinary Journal of Nonlinear Science}\ }\textbf {\bibinfo {volume} {30}},\ \bibinfo {pages} {063133} (\bibinfo {year} {2020})}\BibitemShut {NoStop}%
\bibitem [{\citenamefont {Lu}, \citenamefont {Hunt},\ and\ \citenamefont {Ott}(2018)}]{Lu2018_Attractor_Reconstruction}%
  \BibitemOpen
  \bibfield  {author} {\bibinfo {author} {\bibfnamefont {Z.}~\bibnamefont {Lu}}, \bibinfo {author} {\bibfnamefont {B.~R.}\ \bibnamefont {Hunt}}, \ and\ \bibinfo {author} {\bibfnamefont {E.}~\bibnamefont {Ott}},\ }\bibfield  {title} {\enquote {\bibinfo {title} {{Attractor reconstruction by machine learning}},}\ }\href {\doibase 10.1063/1.5039508} {\bibfield  {journal} {\bibinfo  {journal} {Chaos: An Interdisciplinary Journal of Nonlinear Science}\ }\textbf {\bibinfo {volume} {28}},\ \bibinfo {pages} {061104} (\bibinfo {year} {2018})}\BibitemShut {NoStop}%
\bibitem [{\citenamefont {Patel}\ \emph {et~al.}(2021)\citenamefont {Patel}, \citenamefont {Canaday}, \citenamefont {Girvan}, \citenamefont {Pomerance},\ and\ \citenamefont {Ott}}]{Patel2021_NonstationaryRC}%
  \BibitemOpen
  \bibfield  {author} {\bibinfo {author} {\bibfnamefont {D.}~\bibnamefont {Patel}}, \bibinfo {author} {\bibfnamefont {D.}~\bibnamefont {Canaday}}, \bibinfo {author} {\bibfnamefont {M.}~\bibnamefont {Girvan}}, \bibinfo {author} {\bibfnamefont {A.}~\bibnamefont {Pomerance}}, \ and\ \bibinfo {author} {\bibfnamefont {E.}~\bibnamefont {Ott}},\ }\bibfield  {title} {\enquote {\bibinfo {title} {{Using machine learning to predict statistical properties of non-stationary dynamical processes: System climate,regime transitions, and the effect of stochasticity}},}\ }\href {\doibase 10.1063/5.0042598} {\bibfield  {journal} {\bibinfo  {journal} {Chaos: An Interdisciplinary Journal of Nonlinear Science}\ }\textbf {\bibinfo {volume} {31}},\ \bibinfo {pages} {033149} (\bibinfo {year} {2021})}\BibitemShut {NoStop}%
\bibitem [{\citenamefont {Panahi}\ \emph {et~al.}(2025)\citenamefont {Panahi}, \citenamefont {Kong}, \citenamefont {Glaz}, \citenamefont {Haile},\ and\ \citenamefont {Lai}}]{Panahi2025_CriticalTransitions}%
  \BibitemOpen
  \bibfield  {author} {\bibinfo {author} {\bibfnamefont {S.}~\bibnamefont {Panahi}}, \bibinfo {author} {\bibfnamefont {L.-W.}\ \bibnamefont {Kong}}, \bibinfo {author} {\bibfnamefont {B.}~\bibnamefont {Glaz}}, \bibinfo {author} {\bibfnamefont {M.}~\bibnamefont {Haile}}, \ and\ \bibinfo {author} {\bibfnamefont {Y.-C.}\ \bibnamefont {Lai}},\ }\href {https://arxiv.org/abs/2501.01579} {\enquote {\bibinfo {title} {Unsupervised learning for anticipating critical transitions},}\ } (\bibinfo {year} {2025}),\ \Eprint {http://arxiv.org/abs/2501.01579} {arXiv:2501.01579 [nlin.CD]} \BibitemShut {NoStop}%
\bibitem [{\citenamefont {Jaeger}(2001)}]{Jaeger2001_ESP}%
  \BibitemOpen
  \bibfield  {author} {\bibinfo {author} {\bibfnamefont {H.}~\bibnamefont {Jaeger}},\ }\href {http://www.faculty.jacobs-university.de/hjaeger/pubs/EchoStatesTechRep.pdf} {\enquote {\bibinfo {title} {The "echo state" approach to analysing and training recurrent neural networks},}\ }\bibinfo {type} {GMD Report}\ \bibinfo {number} {148}\ (\bibinfo  {institution} {GMD - German National Research Institute for Computer Science},\ \bibinfo {year} {2001})\BibitemShut {NoStop}%
\bibitem [{\citenamefont {Luko{\v{s}}evi{\v{c}}ius}(2012)}]{Lukosevicius_2012}%
  \BibitemOpen
  \bibfield  {author} {\bibinfo {author} {\bibfnamefont {M.}~\bibnamefont {Luko{\v{s}}evi{\v{c}}ius}},\ }\enquote {\bibinfo {title} {A practical guide to applying echo state networks},}\ in\ \href {\doibase 10.1007/978-3-642-35289-8_36} {\emph {\bibinfo {booktitle} {Neural Networks: Tricks of the Trade: Second Edition}}},\ \bibinfo {editor} {edited by\ \bibinfo {editor} {\bibfnamefont {G.}~\bibnamefont {Montavon}}, \bibinfo {editor} {\bibfnamefont {G.~B.}\ \bibnamefont {Orr}}, \ and\ \bibinfo {editor} {\bibfnamefont {K.-R.}\ \bibnamefont {M{\"u}ller}}}\ (\bibinfo  {publisher} {Springer Berlin Heidelberg},\ \bibinfo {address} {Berlin, Heidelberg},\ \bibinfo {year} {2012})\ pp.\ \bibinfo {pages} {659--686}\BibitemShut {NoStop}%
\bibitem [{\citenamefont {Cucchi}\ \emph {et~al.}(2022)\citenamefont {Cucchi}, \citenamefont {Abreu}, \citenamefont {Ciccone}, \citenamefont {Brunner},\ and\ \citenamefont {Kleemann}}]{Cucchi2022_Hands_On_RC}%
  \BibitemOpen
  \bibfield  {author} {\bibinfo {author} {\bibfnamefont {M.}~\bibnamefont {Cucchi}}, \bibinfo {author} {\bibfnamefont {S.}~\bibnamefont {Abreu}}, \bibinfo {author} {\bibfnamefont {G.}~\bibnamefont {Ciccone}}, \bibinfo {author} {\bibfnamefont {D.}~\bibnamefont {Brunner}}, \ and\ \bibinfo {author} {\bibfnamefont {H.}~\bibnamefont {Kleemann}},\ }\bibfield  {title} {\enquote {\bibinfo {title} {Hands-on reservoir computing: a tutorial for practical implementation},}\ }\href {\doibase 10.1088/2634-4386/ac7db7} {\bibfield  {journal} {\bibinfo  {journal} {Neuromorphic Computing and Engineering}\ }\textbf {\bibinfo {volume} {2}},\ \bibinfo {pages} {032002} (\bibinfo {year} {2022})}\BibitemShut {NoStop}%
\bibitem [{\citenamefont {Platt}\ \emph {et~al.}(2021)\citenamefont {Platt}, \citenamefont {Wong}, \citenamefont {Clark}, \citenamefont {Penny},\ and\ \citenamefont {Abarbanel}}]{Platt2021_PredictiveGS}%
  \BibitemOpen
  \bibfield  {author} {\bibinfo {author} {\bibfnamefont {J.~A.}\ \bibnamefont {Platt}}, \bibinfo {author} {\bibfnamefont {A.}~\bibnamefont {Wong}}, \bibinfo {author} {\bibfnamefont {R.}~\bibnamefont {Clark}}, \bibinfo {author} {\bibfnamefont {S.~G.}\ \bibnamefont {Penny}}, \ and\ \bibinfo {author} {\bibfnamefont {H.~D.~I.}\ \bibnamefont {Abarbanel}},\ }\bibfield  {title} {\enquote {\bibinfo {title} {{Robust forecasting using predictive generalized synchronization in reservoir computing}},}\ }\href {\doibase 10.1063/5.0066013} {\bibfield  {journal} {\bibinfo  {journal} {Chaos: An Interdisciplinary Journal of Nonlinear Science}\ }\textbf {\bibinfo {volume} {31}},\ \bibinfo {pages} {123118} (\bibinfo {year} {2021})}\BibitemShut {NoStop}%
\bibitem [{\citenamefont {Platt}\ \emph {et~al.}(2022)\citenamefont {Platt}, \citenamefont {Penny}, \citenamefont {Smith}, \citenamefont {Chen},\ and\ \citenamefont {Abarbanel}}]{Platt2022_RC_for_Complex_Forecasting_Review}%
  \BibitemOpen
  \bibfield  {author} {\bibinfo {author} {\bibfnamefont {J.~A.}\ \bibnamefont {Platt}}, \bibinfo {author} {\bibfnamefont {S.~G.}\ \bibnamefont {Penny}}, \bibinfo {author} {\bibfnamefont {T.~A.}\ \bibnamefont {Smith}}, \bibinfo {author} {\bibfnamefont {T.-C.}\ \bibnamefont {Chen}}, \ and\ \bibinfo {author} {\bibfnamefont {H.~D.}\ \bibnamefont {Abarbanel}},\ }\bibfield  {title} {\enquote {\bibinfo {title} {A systematic exploration of reservoir computing for forecasting complex spatiotemporal dynamics},}\ }\href {\doibase https://doi.org/10.1016/j.neunet.2022.06.025} {\bibfield  {journal} {\bibinfo  {journal} {Neural Networks}\ }\textbf {\bibinfo {volume} {153}},\ \bibinfo {pages} {530--552} (\bibinfo {year} {2022})}\BibitemShut {NoStop}%
\bibitem [{\citenamefont {Grebogi}\ \emph {et~al.}(1983)\citenamefont {Grebogi}, \citenamefont {McDonald}, \citenamefont {Ott},\ and\ \citenamefont {Yorke}}]{Grebogi1983_FinalStateSensitivity}%
  \BibitemOpen
  \bibfield  {author} {\bibinfo {author} {\bibfnamefont {C.}~\bibnamefont {Grebogi}}, \bibinfo {author} {\bibfnamefont {S.~W.}\ \bibnamefont {McDonald}}, \bibinfo {author} {\bibfnamefont {E.}~\bibnamefont {Ott}}, \ and\ \bibinfo {author} {\bibfnamefont {J.~A.}\ \bibnamefont {Yorke}},\ }\bibfield  {title} {\enquote {\bibinfo {title} {Final state sensitivity: An obstruction to predictability},}\ }\href {\doibase https://doi.org/10.1016/0375-9601(83)90945-3} {\bibfield  {journal} {\bibinfo  {journal} {Physics Letters A}\ }\textbf {\bibinfo {volume} {99}},\ \bibinfo {pages} {415--418} (\bibinfo {year} {1983})}\BibitemShut {NoStop}%
\bibitem [{\citenamefont {Canaday}\ \emph {et~al.}(2024)\citenamefont {Canaday}, \citenamefont {Kalra}, \citenamefont {Wikner}, \citenamefont {Norton}, \citenamefont {Hunt},\ and\ \citenamefont {Pomerance}}]{rescompy}%
  \BibitemOpen
  \bibfield  {author} {\bibinfo {author} {\bibfnamefont {D.}~\bibnamefont {Canaday}}, \bibinfo {author} {\bibfnamefont {D.}~\bibnamefont {Kalra}}, \bibinfo {author} {\bibfnamefont {A.}~\bibnamefont {Wikner}}, \bibinfo {author} {\bibfnamefont {D.~A.}\ \bibnamefont {Norton}}, \bibinfo {author} {\bibfnamefont {B.}~\bibnamefont {Hunt}}, \ and\ \bibinfo {author} {\bibfnamefont {A.}~\bibnamefont {Pomerance}},\ }\bibfield  {title} {\enquote {\bibinfo {title} {{{rescompy} 1.0.0: Fundamental Methods for Reservoir Computing in Python}},}\ }\href@noop {} {\bibfield  {journal} {\bibinfo  {journal} {GitHub}\ } (\bibinfo {year} {2024})}\BibitemShut {NoStop}%
\bibitem [{\citenamefont {Wikner}\ \emph {et~al.}(2024)\citenamefont {Wikner}, \citenamefont {Harvey}, \citenamefont {Girvan}, \citenamefont {Hunt}, \citenamefont {Pomerance}, \citenamefont {Antonsen},\ and\ \citenamefont {Ott}}]{Wikner2024_LMNT}%
  \BibitemOpen
  \bibfield  {author} {\bibinfo {author} {\bibfnamefont {A.}~\bibnamefont {Wikner}}, \bibinfo {author} {\bibfnamefont {J.}~\bibnamefont {Harvey}}, \bibinfo {author} {\bibfnamefont {M.}~\bibnamefont {Girvan}}, \bibinfo {author} {\bibfnamefont {B.~R.}\ \bibnamefont {Hunt}}, \bibinfo {author} {\bibfnamefont {A.}~\bibnamefont {Pomerance}}, \bibinfo {author} {\bibfnamefont {T.}~\bibnamefont {Antonsen}}, \ and\ \bibinfo {author} {\bibfnamefont {E.}~\bibnamefont {Ott}},\ }\bibfield  {title} {\enquote {\bibinfo {title} {Stabilizing machine learning prediction of dynamics: Novel noise-inspired regularization tested with reservoir computing},}\ }\href {\doibase https://doi.org/10.1016/j.neunet.2023.10.054} {\bibfield  {journal} {\bibinfo  {journal} {Neural Networks}\ }\textbf {\bibinfo {volume} {170}},\ \bibinfo {pages} {94--110} (\bibinfo {year} {2024})}\BibitemShut {NoStop}%
\bibitem [{\citenamefont {Tikhonov}\ \emph {et~al.}(1995)\citenamefont {Tikhonov}, \citenamefont {Goncharsky}, \citenamefont {Stepanov},\ and\ \citenamefont {Yagola}}]{Tikhonov1995}%
  \BibitemOpen
  \bibfield  {author} {\bibinfo {author} {\bibfnamefont {A.~N.}\ \bibnamefont {Tikhonov}}, \bibinfo {author} {\bibfnamefont {A.~V.}\ \bibnamefont {Goncharsky}}, \bibinfo {author} {\bibfnamefont {V.~V.}\ \bibnamefont {Stepanov}}, \ and\ \bibinfo {author} {\bibfnamefont {A.~G.}\ \bibnamefont {Yagola}},\ }\enquote {\bibinfo {title} {Regularization methods},}\ in\ \href {\doibase 10.1007/978-94-015-8480-7_2} {\emph {\bibinfo {booktitle} {Numerical Methods for the Solution of Ill-Posed Problems}}}\ (\bibinfo  {publisher} {Springer Netherlands},\ \bibinfo {address} {Dordrecht},\ \bibinfo {year} {1995})\ pp.\ \bibinfo {pages} {7--63}\BibitemShut {NoStop}%
\bibitem [{\citenamefont {Grigoryeva}\ \emph {et~al.}(2024)\citenamefont {Grigoryeva}, \citenamefont {Hamzi}, \citenamefont {Kemeth}, \citenamefont {Kevrekidis}, \citenamefont {Manjunath}, \citenamefont {Ortega},\ and\ \citenamefont {Steynberg}}]{Grigoryeva2024_ColdStartRC}%
  \BibitemOpen
  \bibfield  {author} {\bibinfo {author} {\bibfnamefont {L.}~\bibnamefont {Grigoryeva}}, \bibinfo {author} {\bibfnamefont {B.}~\bibnamefont {Hamzi}}, \bibinfo {author} {\bibfnamefont {F.~P.}\ \bibnamefont {Kemeth}}, \bibinfo {author} {\bibfnamefont {Y.}~\bibnamefont {Kevrekidis}}, \bibinfo {author} {\bibfnamefont {G.}~\bibnamefont {Manjunath}}, \bibinfo {author} {\bibfnamefont {J.-P.}\ \bibnamefont {Ortega}}, \ and\ \bibinfo {author} {\bibfnamefont {M.~J.}\ \bibnamefont {Steynberg}},\ }\bibfield  {title} {\enquote {\bibinfo {title} {Data-driven cold starting of good reservoirs},}\ }\href {\doibase https://doi.org/10.1016/j.physd.2024.134325} {\bibfield  {journal} {\bibinfo  {journal} {Physica D: Nonlinear Phenomena}\ }\textbf {\bibinfo {volume} {469}},\ \bibinfo {pages} {134325} (\bibinfo {year} {2024})}\BibitemShut {NoStop}%
\bibitem [{\citenamefont {Duffing}(1918)}]{Duffing_1918}%
  \BibitemOpen
  \bibfield  {author} {\bibinfo {author} {\bibfnamefont {G.}~\bibnamefont {Duffing}},\ }\href@noop {} {\emph {\bibinfo {title} {Erzwungene Schwingungen Bei Ver\"aNderlicher Eigenfrequenz Und Ihre Technische Bedeutung}}},\ \bibinfo {number} {41-41}\ (\bibinfo  {publisher} {F. Vieweg \& Sohn},\ \bibinfo {year} {1918})\BibitemShut {NoStop}%
\bibitem [{\citenamefont {Motter}\ \emph {et~al.}(2013)\citenamefont {Motter}, \citenamefont {Gruiz}, \citenamefont {K\'arolyi},\ and\ \citenamefont {T\'el}}]{Motter2013_DoublyTransient}%
  \BibitemOpen
  \bibfield  {author} {\bibinfo {author} {\bibfnamefont {A.~E.}\ \bibnamefont {Motter}}, \bibinfo {author} {\bibfnamefont {M.}~\bibnamefont {Gruiz}}, \bibinfo {author} {\bibfnamefont {G.}~\bibnamefont {K\'arolyi}}, \ and\ \bibinfo {author} {\bibfnamefont {T.}~\bibnamefont {T\'el}},\ }\bibfield  {title} {\enquote {\bibinfo {title} {Doubly transient chaos: Generic form of chaos in autonomous dissipative systems},}\ }\href {\doibase 10.1103/PhysRevLett.111.194101} {\bibfield  {journal} {\bibinfo  {journal} {Phys. Rev. Lett.}\ }\textbf {\bibinfo {volume} {111}},\ \bibinfo {pages} {194101} (\bibinfo {year} {2013})}\BibitemShut {NoStop}%
\bibitem [{\citenamefont {L\"{u}}, \citenamefont {Chen},\ and\ \citenamefont {Cheng}(2004)}]{Lu2004_MultistableLorenz}%
  \BibitemOpen
  \bibfield  {author} {\bibinfo {author} {\bibfnamefont {J.}~\bibnamefont {L\"{u}}}, \bibinfo {author} {\bibfnamefont {G.}~\bibnamefont {Chen}}, \ and\ \bibinfo {author} {\bibfnamefont {D.}~\bibnamefont {Cheng}},\ }\bibfield  {title} {\enquote {\bibinfo {title} {A new chaotic system and beyond: The generalized lorenz-like system},}\ }\href {\doibase 10.1142/S021812740401014X} {\bibfield  {journal} {\bibinfo  {journal} {International Journal of Bifurcation and Chaos}\ }\textbf {\bibinfo {volume} {14}},\ \bibinfo {pages} {1507--1537} (\bibinfo {year} {2004})}\BibitemShut {NoStop}%
\bibitem [{\citenamefont {Chen}, \citenamefont {Nishikawa},\ and\ \citenamefont {Motter}(2017)}]{Chen2017_SlimFractals}%
  \BibitemOpen
  \bibfield  {author} {\bibinfo {author} {\bibfnamefont {X.}~\bibnamefont {Chen}}, \bibinfo {author} {\bibfnamefont {T.}~\bibnamefont {Nishikawa}}, \ and\ \bibinfo {author} {\bibfnamefont {A.~E.}\ \bibnamefont {Motter}},\ }\bibfield  {title} {\enquote {\bibinfo {title} {Slim fractals: The geometry of doubly transient chaos},}\ }\href {\doibase 10.1103/PhysRevX.7.021040} {\bibfield  {journal} {\bibinfo  {journal} {Phys. Rev. X}\ }\textbf {\bibinfo {volume} {7}},\ \bibinfo {pages} {021040} (\bibinfo {year} {2017})}\BibitemShut {NoStop}%
\bibitem [{\citenamefont {Zhang}\ and\ \citenamefont {Gilpin}(2025)}]{Zhang2025_Zeroshot}%
  \BibitemOpen
  \bibfield  {author} {\bibinfo {author} {\bibfnamefont {Y.}~\bibnamefont {Zhang}}\ and\ \bibinfo {author} {\bibfnamefont {W.}~\bibnamefont {Gilpin}},\ }\bibfield  {title} {\enquote {\bibinfo {title} {Zero-shot forecasting of chaotic systems},}\ }in\ \href {https://openreview.net/forum?id=TqYjhJrp9m} {\emph {\bibinfo {booktitle} {The Thirteenth International Conference on Learning Representations}}}\ (\bibinfo {year} {2025})\BibitemShut {NoStop}%
\bibitem [{\citenamefont {Hess}\ \emph {et~al.}(2023)\citenamefont {Hess}, \citenamefont {Monfared}, \citenamefont {Brenner},\ and\ \citenamefont {Durstewitz}}]{Hess2023_TeachingForcing}%
  \BibitemOpen
  \bibfield  {author} {\bibinfo {author} {\bibfnamefont {F.}~\bibnamefont {Hess}}, \bibinfo {author} {\bibfnamefont {Z.}~\bibnamefont {Monfared}}, \bibinfo {author} {\bibfnamefont {M.}~\bibnamefont {Brenner}}, \ and\ \bibinfo {author} {\bibfnamefont {D.}~\bibnamefont {Durstewitz}},\ }\bibfield  {title} {\enquote {\bibinfo {title} {Generalized teacher forcing for learning chaotic dynamics},}\ }in\ \href@noop {} {\emph {\bibinfo {booktitle} {Proceedings of the 40th International Conference on Machine Learning}}},\ \bibinfo {series and number} {ICML'23}\ (\bibinfo  {publisher} {JMLR.org},\ \bibinfo {year} {2023})\BibitemShut {NoStop}%
\bibitem [{\citenamefont {Feng}\ and\ \citenamefont {Tu}(2021)}]{Feng2021_FlatMinima}%
  \BibitemOpen
  \bibfield  {author} {\bibinfo {author} {\bibfnamefont {Y.}~\bibnamefont {Feng}}\ and\ \bibinfo {author} {\bibfnamefont {Y.}~\bibnamefont {Tu}},\ }\bibfield  {title} {\enquote {\bibinfo {title} {The inverse variance–flatness relation in stochastic gradient descent is critical for finding flat minima},}\ }\href {\doibase 10.1073/pnas.2015617118} {\bibfield  {journal} {\bibinfo  {journal} {Proceedings of the National Academy of Sciences}\ }\textbf {\bibinfo {volume} {118}},\ \bibinfo {pages} {e2015617118} (\bibinfo {year} {2021})}\BibitemShut {NoStop}%
\bibitem [{\citenamefont {Belkin}\ \emph {et~al.}(2019)\citenamefont {Belkin}, \citenamefont {Hsu}, \citenamefont {Ma},\ and\ \citenamefont {Mandal}}]{Belkin2019_DoubleDescent}%
  \BibitemOpen
  \bibfield  {author} {\bibinfo {author} {\bibfnamefont {M.}~\bibnamefont {Belkin}}, \bibinfo {author} {\bibfnamefont {D.}~\bibnamefont {Hsu}}, \bibinfo {author} {\bibfnamefont {S.}~\bibnamefont {Ma}}, \ and\ \bibinfo {author} {\bibfnamefont {S.}~\bibnamefont {Mandal}},\ }\bibfield  {title} {\enquote {\bibinfo {title} {Reconciling modern machine-learning practice and the classical bias–variance trade-off},}\ }\href {\doibase 10.1073/pnas.1903070116} {\bibfield  {journal} {\bibinfo  {journal} {Proceedings of the National Academy of Sciences}\ }\textbf {\bibinfo {volume} {116}},\ \bibinfo {pages} {15849--15854} (\bibinfo {year} {2019})}\BibitemShut {NoStop}%
\bibitem [{\citenamefont {Gilpin}(2021)}]{Gilpin2021_ChaosBenchmark}%
  \BibitemOpen
  \bibfield  {author} {\bibinfo {author} {\bibfnamefont {W.}~\bibnamefont {Gilpin}},\ }\bibfield  {title} {\enquote {\bibinfo {title} {Chaos as an interpretable benchmark for forecasting and data-driven modelling},}\ }in\ \href@noop {} {\emph {\bibinfo {booktitle} {Advances in Neural Information Processing}}}\ (\bibinfo {year} {2021})\BibitemShut {NoStop}%
\bibitem [{\citenamefont {Gilpin}(2023)}]{Gilpin2023_ModelScaleDomainKnowledge}%
  \BibitemOpen
  \bibfield  {author} {\bibinfo {author} {\bibfnamefont {W.}~\bibnamefont {Gilpin}},\ }\bibfield  {title} {\enquote {\bibinfo {title} {Model scale versus domain knowledge in statistical forecasting of chaotic systems},}\ }\href {\doibase 10.1103/PhysRevResearch.5.043252} {\bibfield  {journal} {\bibinfo  {journal} {Phys. Rev. Res.}\ }\textbf {\bibinfo {volume} {5}},\ \bibinfo {pages} {043252} (\bibinfo {year} {2023})}\BibitemShut {NoStop}%
\end{thebibliography}%
